\documentclass[10pt,journal,compsoc]{IEEEtran}
\newif\ifpeerreview

\peerreviewfalse

\usepackage[nocompress]{cite}
\usepackage{url}
\usepackage{amsmath,amssymb,graphicx}

\usepackage{lipsum} 

\usepackage[switch]{lineno}

\usepackage{hyperref}
\usepackage{amsmath,amsfonts}
\usepackage{algorithmic}
\usepackage{algorithm}
\usepackage{array}
\usepackage{textcomp}
\usepackage{stfloats}
\usepackage{url}
\usepackage{verbatim}
\usepackage{graphicx}
\usepackage{booktabs}  
\usepackage{colortbl}
\usepackage{romannum}
\usepackage{amsthm}
\usepackage{arydshln}
\usepackage{microtype}
\usepackage{rotating}
\usepackage[capitalize]{cleveref}
\usepackage{xcolor}
\usepackage{pifont}
\usepackage{colortbl} 
\newcommand{\cmark}{\ding{51}} 
\newcommand{\xmark}{\ding{55}} 
\definecolor{lightcyan}{rgb}{0.88, 1.0, 1.0}

\newcommand{\etal}{\textit{et al.}}
\newcommand{\best}[1]{\textbf{#1}}
\newcommand{\second}[1]{\underline{#1}}

\newcommand{\liying}[1]{{\color{black}{#1}}}
\newcommand{\pami}[1]{{\color{black}{#1}}}

\newtheorem{proposition}{Proposition}

\usepackage{multirow,multicol,xspace}
\usepackage{adjustbox}
\makeatletter
\def\adl@drawiv#1#2#3{%
        \hskip.5\tabcolsep
        \xleaders#3{#2.5\@tempdimb #1{1}#2.5\@tempdimb}%
                #2\z@ plus1fil minus1fil\relax
        \hskip.5\tabcolsep}
\newcommand{\cdashlinelr}[1]{%
  \noalign{\vskip\aboverulesep
           \global\let\@dashdrawstore\adl@draw
           \global\let\adl@draw\adl@drawiv}
  \cdashline{#1}
  \noalign{\global\let\adl@draw\@dashdrawstore
           \vskip\belowrulesep}}
\makeatother

\newcommand{\paperID}{10}

\title{Dark Noise Diffusion: Noise Synthesis for Low-Light Image Denoising}

\author{Liying Lu,~\IEEEmembership{Student Member,~IEEE,}
        Raphaël Achddou,~\IEEEmembership{Member,~IEEE,}
        and~Sabine Süsstrunk,~\IEEEmembership{Fellow,~IEEE}
\IEEEcompsocitemizethanks{\IEEEcompsocthanksitem All authors are with the School of Computer and Communication Sciences, École Polytechnique Fédérale de Lausanne, Lausanne,
VD, 1015.\protect\\
E-mail: \{liying.lu, raphael.achddou, sabine.susstrunk\}@epfl.ch
}
}

\begin{document}

\IEEEtitleabstractindextext{%
\begin{abstract}
Low-light photography produces images with low signal-to-noise ratios due to limited photons. In such conditions, common approximations like the Gaussian noise model fall short, and many denoising techniques fail to remove noise effectively. Although deep-learning methods perform well, they require large datasets of paired images that are impractical to acquire. As a remedy, synthesizing realistic low-light noise has gained significant attention. In this paper, we investigate the ability of diffusion models to capture the complex distribution of low-light noise. We show that a naive application of conventional diffusion models is inadequate for this task and propose three key adaptations that enable high-precision noise generation: a two-branch architecture to better model signal-dependent and signal-independent noise, the incorporation of positional information to capture fixed-pattern noise, and a tailored diffusion noise schedule. Consequently, our model enables the generation of large datasets for training low-light denoising networks, leading to state-of-the-art performance. Through comprehensive analysis, including statistical evaluation and noise decomposition, we provide deeper insights into the characteristics of the generated data.

\end{abstract}

}

\maketitle





\IEEEraisesectionheading{
  \section{Introduction}\label{sec:intro}
}

\IEEEPARstart{O}{btaining} noiseless photographs in low-light settings is extremely challenging.
Although extending the exposure time enhances brightness by gathering more light, this solution falls short for dynamic scenes due to the necessity of a static environment during image capture. 
Alternatively, increasing the system gain expands the RAW signal's dynamic range, but this operation also dramatically increases the noise level in the captured image.

To address this, numerous deep-learning denoising methods~\cite{zhang2017beyond,mao2016image,tai2017memnet} have been proposed. However, training deep networks requires substantial amounts of data, and acquiring paired noisy-clean image sets is both laborious and time-consuming. 
This process involves capturing a clean image with a long-exposure time and a noisy image with a short-exposure time under the same low-light condition. To ensure proper alignment, the camera must remain stationary and the scene must be free of moving objects. These strict conditions make data collection difficult and limit the dataset to static scenes.


\begin{figure}[htp]
  \centering
  \includegraphics[width=1.0\columnwidth]{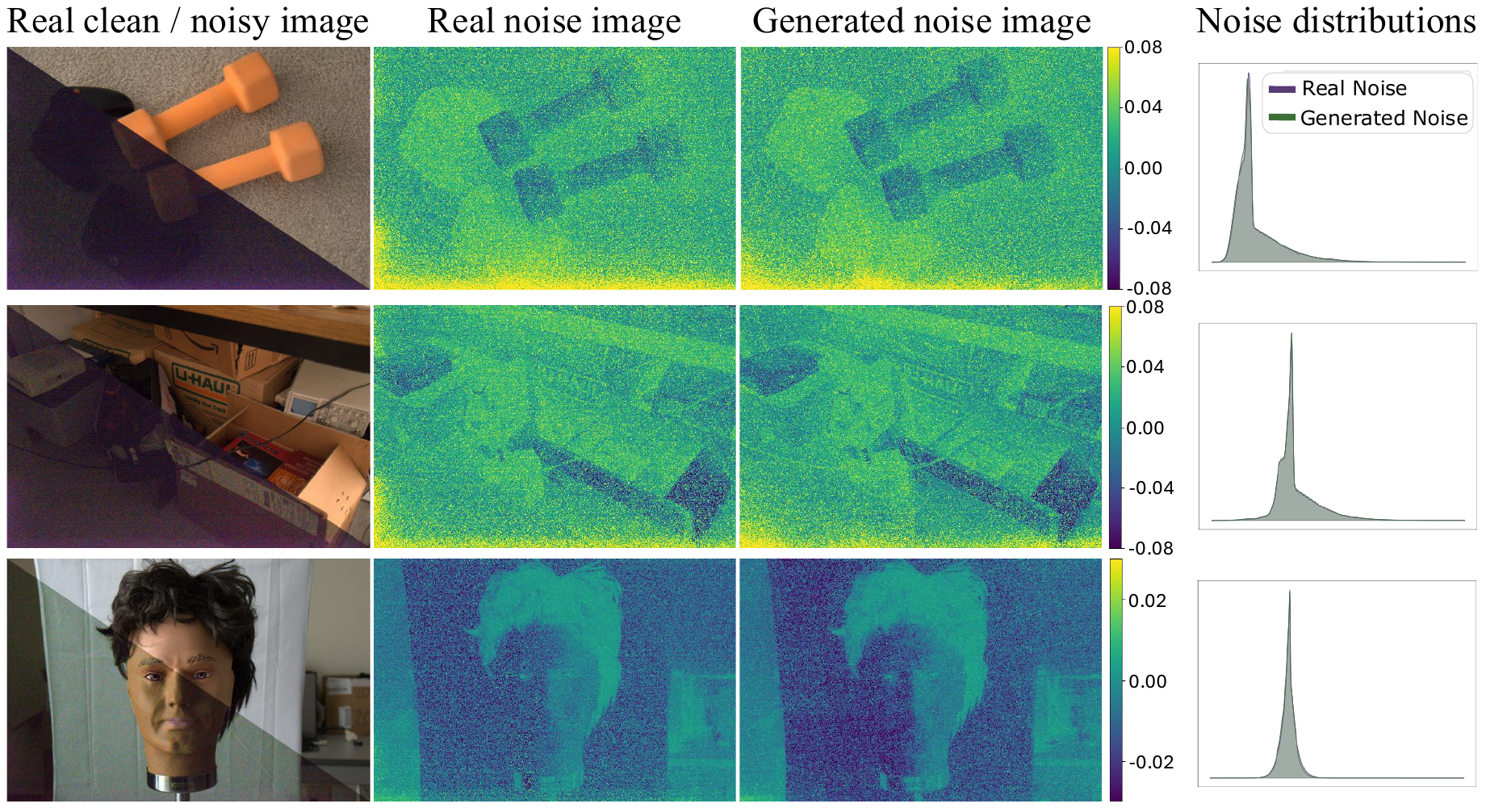}
  \caption{
  Examples of noise images generated by our Dark \textbf{Noise Diff}usion~(NoiseDiff) method for three images from the SID Sony dataset~\cite{chen2018learning}. Noise values are represented following a color map displayed on the right of the generated noise images.  The generated noise images closely resemble real noise images. The distribution of the generated noise also aligns well with the real noise distributions.
  }
  \label{fig:teaser}
\end{figure}

As larger training datasets generally lead to better low-light denoising performance, the challenge of obtaining sufficient training data has become a prominent topic in the field.
To address this bottleneck, a growing number of methods have embraced noise synthesis techniques that can be broadly categorized into physics-based and deep-learning based methods.
In physics-based noise modeling, the prevalent Poisson-Gaussian noise model combines signal-dependent Poisson and signal-independent Gaussian distributions. 
Despite advances in physics-based noise modeling by subsequent works \cite{wach2004noise,costantini2004virtual,zhang2017improved,foi2009clipped,wei2021physics}, a disparity persists between the synthesized and real noise distributions.
Conversely, deep-learning-based noise modeling explores the application of generative neural networks for learning the intricate patterns of real-world noise, either using generative adversarial networks (GANs)~\cite{jang2021c2n,monakhova2022dancing,zhang2023towards} or normalizing flows~\cite{abdelhamed2019noise}.
However, GANs generally suffer from training instability and struggle to learn multimodal distributions effectively~\cite{salmona2022can}. 
Normalizing flows are limited in model architecture design and also face difficulties with highly multimodal distributions~\cite{bond2021deep}.

In contrast, diffusion models~\cite{ho2020denoising,dhariwal2021diffusion,rombach2022high} have shown strong performance in capturing complex multi-modal data distributions.
In this paper, we investigate their application to the task of low-light noise synthesis. 
While a direct application of conventional diffusion models proves inadequate for this task, our proposed framework, Dark \textbf{Noise Diff}usion~(NoiseDiff), achieves state-of-the-art noise synthesis performance thanks to several adaptations, regarding the architecture of the diffusion network, and the diffusion noise scheduling scheme. 
As illustrated in \cref{fig:teaser}, our method generates noise with high precision, effectively capturing both local and global features.
For each proposed adaptation, we investigate its relevance by carefully analyzing various statistics of the generated noise in Sec.~\ref{sec:discussions}.
Our contributions are as follows: (1) We demonstrate that a two-branch neural network is an effective approach for modeling signal-dependent and signal-independent noise. (2) We incorporate positional encoding to account for the spatially varying nature of signal-independent noise. (3) We show empirically and theoretically that the diffusion noise schedule is critical for accurately capturing the noise variance. (4) We obtain state-of-the-art low-light noise synthesis performance, leading to better denoising performance compared to models trained on data synthesized by other techniques.

\section{Related Work}\label{sec:rel_works}


\noindent \textbf{Image Denoising.}
Image denoising remains a persistent challenge in computer vision and image processing. 
Traditional methods often rely on various image priors, such as sparsity~\cite{portilla2003image,elad2006image,aharon2006k,mairal2009non,mairal2007sparse,xie2017kronecker,xu2018trilateral,zhou2009non}, self-similarity~\cite{buades2005non,dabov2007image}, and smoothness~\cite{rudin1992nonlinear}. 
Notably, BM3D~\cite{dabov2007image} is still regarded as a standard approach due to its accuracy and robustness.

Recent advancements in deep learning have revolutionized the field of image denoising, with Convolutional Neural Networks (CNNs) surpassing traditional methods to set new benchmarks~\cite{zhang2017beyond,mao2016image,tai2017memnet,gharbi2016deep,zamir2022restormer}. However, most of these works approximate the noise distribution as Gaussian, whereas real noise distributions are far more complex. More recent works~\cite{abdelhamed2018high,anaya2018renoir,chen2018learning,wang2020practical,flepp2024real} focus on using real-world datasets to train denoising models.

Although image denoising datasets are available \cite{abdelhamed2018high,anaya2018renoir,chen2018learning}, they have several limitations: (1) data quantity is limited due to the tedious process of acquiring real-world datasets of clean and noisy image pairs. (2) Samples often have similar content as they are captured in limited scenes. (3) They are restricted to still scenes, as clean and noisy images must be well-aligned.
These limitations impede image-denoising models from attaining optimal performance, as effectively training deep neural networks requires extensive training data.
Consequently, numerous studies have begun exploring realistic noise synthesis as an easier alternative to data acquisition for increasing data quantity.

\noindent \textbf{Physics-Based Noise Modeling.}
The widely used Additive White Gaussian Noise (AWGN) approximation can not model real-world noise correctly.
A more realistic alternative is the Poisson-Gaussian noise model~\cite{foi2008practical,foi2009clipped,brooks2019unprocessing,bahler2022pogain}, which better reflects the nature of noise in imaging sensors.
This model decomposes noise into two components: a signal-dependent Poisson noise arising from the stochastic nature of photon capture, and a signal-independent Gaussian noise approximating sensor-readout and thermal noises. 

\liying{
During photon collection, let $u^*$ denote the expected number of incident photons, $\alpha$ the quantum efficiency of the sensor, and $g$ the system gain of the analog amplifier, commonly referred to as ISO. The ideal signal, assuming perfect photon detection, is then given by: $x^{*} = g \alpha u^{*} \label{eq:clean}$. However, the actual number of collected photons, $u$, is modeled as a Poisson distribution with mean $u^*$: 
$u \sim \mathcal{P}(u^{*})$. 
In addition to photon shot noise, the sensor introduces dark current noise, $n_d$, which is usually modeled as a Gaussian distribution: $n_d \sim \mathcal{N}(0, \sigma_{d}^2)$.

Subsequently, during the signal readout process, additional signal-independent noise sources, including read noise, are introduced. These are commonly modeled as a zero-mean Gaussian distribution $n_r \sim \mathcal{N}(0, \sigma_{r}^2)$.
The actual observed signal $x$ can be formulated as
\begin{equation}
  \scalebox{1.0}{$
    \begin{aligned}
      x = g (\alpha u + n_d) + n_r = g\alpha \mathcal{P}(u^{*}) + \mathcal{N}(0, \sigma^2) \label{eq:poisson_gaussian} \;,
    \end{aligned}
  $}
\end{equation}
where $\sigma^2=g^{2}\sigma_{d}^2+\sigma_{r}^2$. 
The noise component, defined as the deviation from the ideal signal, is then
\begin{align}
\label{eq:noise}
    n = x - x^* = g\alpha \mathcal{P}(u^{*}) - g \alpha u^{*} + \mathcal{N}(0, \sigma^2) \;.
\end{align}
}

However, this model remains unable to capture all relevant noise sources. 
Subsequent works~\cite{wach2004noise,costantini2004virtual,wei2021physics,zhang2017improved,foi2009clipped} introduce more complex and accurate noise models.
In addition to the previously mentioned noise sources, these models incorporate other components of real-world noise, such as fixed-pattern noise and color-biased read noise. 
Recently, the ELD paper~\cite{wei2021physics} introduces a noise model that accounts for other noise sources. 
After calibration, this noise model was utilized to synthesize noisy data for training an image-denoising CNN.
Despite these advancements in physical noise modeling, a discrepancy persists between synthetic and real-world noise. 
In particular, modeling the signal-independent component of real noise is extremely challenging.

To address this, Zhang~\etal~\cite{zhang2021rethinking} propose to capture frames in a completely dark environment and sample pixel values from them to represent the signal-independent noise.
Signal-dependent noise is modeled using a simple Poisson distribution.
Similarly, Feng~\etal~\cite{feng2022learnability} further propose a \textit{"dark shading"} correction method utilizing these dark frames to reduce the complexity of real-world noise distribution. 
They capture hundreds of dark frames and compute the average as the \textit{dark shading} at various ISOs, which represents a combination of black level error (BLE) \cite{nakamura2017image} and fixed-pattern noise (FPN) \cite{cain2001projection,holst2007cmos}.
Subtracting this \textit{dark shading} from noisy images eases the denoising process. 
Several subsequent works employ this approach to simplify noise modeling~\cite{cao2023physics} or improve denoising performance~\cite{wang2023exposurediffusion}. 
However, dark frames are not always accessible. In this work, we assume such data is not available.

 \noindent \textbf{Deep Learning-Based Noise Modeling.}
Another research direction focuses on using deep neural networks to capture real-world noise more precisely. 
C2N~\cite{jang2021c2n} proposes to train a generative adversarial network (GAN)~\cite{goodfellow2014generative} with unpaired noisy and clean images for noisy image generation.
The Noise Flow method~\cite{abdelhamed2019noise} introduces the use of normalizing flows~\cite{dinh2016density,kingma2018glow,kobyzev2020normalizing} to model real-world RAW noise. 
\cite{kousha2022modeling} also adopts normalizing flows for noise modeling, but in the output sRGB space.

Recently, LRD \cite{zhang2023towards} proposes modeling signal-dependent noise with a Poisson distribution and learning the signal-independent part with a GAN. The GAN loss enables synthesizing signal-independent noise without explicitly decomposing the full noise. Monakhova et al. \cite{monakhova2022dancing} introduced a noise generator where statistical noise parameters of a physics-based model are learned instead of manually calibrated.

Despite these advances, these methods suffer from several limitations.
Normalizing flows struggle to capture multimodal or intricate structures in the data, and performance is sensitive to the choice of flow architecture, hence requiring careful tuning and experimentation~\cite{bond2021deep}.
As for GANs, they are notorious for their unstable training process~\cite{gulrajani2017improved,mescheder2018training}. 
Diffusion models have demonstrated strong capabilities in modeling complex data distributions~\cite{ramesh2022hierarchical,rombach2022high,poole2022dreamfusion}. 
Therefore, in this paper, we investigate the use of diffusion models to model low-light noise to enable the generation of more realistic data samples.

\section{Method}
\label{sec:method}
\liying{
Conventional diffusion models for image generation cannot be directly applied to low-light noise generation. 
We propose several crucial adaptations that enable high-precision noise generation, regarding (1) the architecture of the diffusion network, (2) the conditioning of our generation algorithm, and (3) the noise schedule of the diffusion process. }
(Please refer to the appendix for preliminaries on diffusion models.)


\begin{figure*}[t]
  \centering
  \includegraphics[width=1\linewidth]{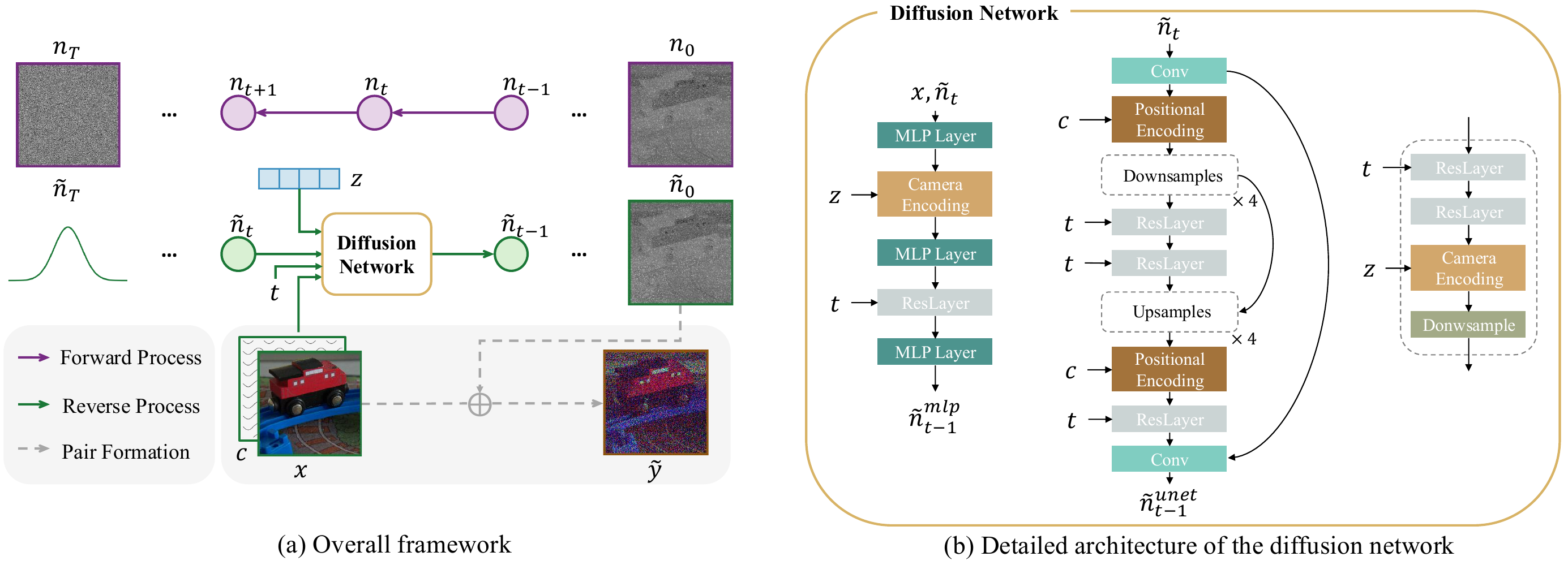}
  \caption{
  (a) The proposed \liying{Dark Noise Diffusion~(NoiseDiff) framework} for synthetic data generation. 
  The forward process progressively destroys a real low-light noise $\mathbf{n}_0$ by adding Gaussian noise. 
  In the reverse process, $\Tilde{\mathbf{n}}_T$ is sampled from a Gaussian distribution and progressively denoised by a network back to the low-light noise $\Tilde{\mathbf{n}}_{0}$. 
  The network is conditioned on the clean image $\mathbf{x}$, pixel coordinates $\mathbf{c}$, and camera settings $\mathbf{z}$, as all these factors influence the noise pattern. 
  Once the low-light noise $\Tilde{\mathbf{n}}_{0}$ is obtained, it can be added to the clean image $\mathbf{x}$ to create the corresponding noisy image $\Tilde{\mathbf{y}}$, thus forming a synthetic clean-noisy pair $(\mathbf{x}, \Tilde{\mathbf{y}})$. 
  (b) The detailed architecture of the two-branch diffusion network, shown from left to right: the MLP branch, the UNet branch, and the architecture of the down-sampling block (with up-sampling blocks having a similar structure).
  }
  \label{fig:network}
\end{figure*}

\subsection{Framework for Low-light Noise Generation}
\label{sec:overall_framework}

Our model is trained on real noisy-clean RAW image pairs from the SID dataset~\cite{chen2018learning}. The noisy image is captured with a short exposure time ($\sim$0.1~s), while the clean image is captured with a long exposure time ($\sim$10~s). In the following sections, we refer to the ratio between these exposure times as the \textit{exposure ratio}.

The overall \liying{NoiseDiff} framework is depicted in Fig.~\ref{fig:network} (a). 
The initial input for the forward process of the diffusion model is a low-light noise image  $\mathbf{n}_0$, obtained by subtracting a clean image from its noisy counterpart (the latter is obtained by multiplying the low-exposure image by the exposure ratio to match the dynamic range of the clean image).
Random samples of Gaussian noise are then added to $\mathbf{n}_0$, gradually approaching a standard normal distribution after $T$ steps. 
For the backward process, a network is employed to predict the added Gaussian noise and denoise $\Tilde{\mathbf{n}}_T$ back to $\Tilde{\mathbf{n}}_{0}$.

As signal-dependent noise is correlated to the clean image and camera settings, we condition our diffusion network on these factors. 
We also encode pixel positions into the diffusion network to better capture spatially related noise components, such as fixed-pattern noise, which is salient in low-light images.
Hence, the reverse process of our diffusion model is formulated as $\Tilde{\mathbf{n}}_{t} = f_{\theta} (\Tilde{\mathbf{n}}_{t-1}, t, \mathbf{x}, \mathbf{c}, \mathbf{z})$, where $t$ is the diffusion timestep, $\mathbf{x}$ is the clean image, $\mathbf{c}$ are the pixel coordinates, and $\mathbf{z}$ is the camera setting (including ISO value and the exposure ratio). 
Conditioning the diffusion model on the clean image is done by concatenating it with the input noise image. 
Regarding the encoding of the pixel coordinates, the camera settings, and the diffusion timestep, further explanation is provided in Sec.~\ref{sec:architecture}.

\subsection{Diffusion Network Architecture}
\label{sec:architecture}

\noindent \textbf{Two-Branch Architecture.}
We design our diffusion network with two branches as shown in Fig.~\ref{fig:network} (b): a Multilayer perceptron (MLP) branch and a UNet branch.
This design addresses the distinct nature of different noise components. 
 Noise components such as shot noise only depend on the current clean pixel value.
Hence, their generation should not be influenced by neighboring pixels, making the MLP branch well-suited for handling these characteristics.
Conversely, noise components that are locally correlated are better captured by the UNet architecture that excels at modeling local dependencies.
\liying{It is important to note that we do not explicitly enforce the MLP branch to model only shot noise or the UNet branch to model other noise components, as the ground-truth separation of these noise types is unavailable. However, as will be demonstrated in the experiment section, this design improves noise modeling accuracy.}

The MLP branch consists of three MLP layers, a camera setting encoding layer and a residual layer with diffusion timestep encoding. 
The UNet branch includes four down-sampling blocks and four up-sampling blocks. 
The UNet branch also includes a camera setting encoding layer, positional encoding layers, and diffusion timestep encoding layers. 
Next, we detail how these additional layers work.

 \noindent \textbf{Positional Encoding.}
As previously stated, signal-independent noise cannot be accurately represented by a Gaussian distribution alone.
Low-light noise includes spatially correlated noise components, such as banding patterns and low-frequency noise \cite{wei2021physics,feng2022learnability}.
Fig.~\ref{fig:noise_visualization} presents two examples from the SID dataset.
Banding patterns and a spatially varying offset at the bottom of the frame are visible, especially in the first example.
We therefore integrate positional encoding in our diffusion network, so that the diffusion model becomes aware of pixel positions.
We first use sinusoidal positional encoding~\cite{vaswani2017attention} to encode the pixel coordinate map $\mathbf{c}$, and use $1\times1$ convolutional layers to learn a scale and shift map $\mathbf{s}$, $\mathbf{r}$, which are then integrated into the image feature as: 
$\Tilde{\mathbf{f}} = \mathbf{f} \odot (1+\mathbf{\mathbf{s}}) + \mathbf{\mathbf{r}}$.
We encode the diffusion timestep $t$ in a similar fashion to that of positional encoding.
For more details, please refer to the appendix.
 
\noindent \textbf{Camera Setting Encoding.}
As camera settings (such as ISO value and exposure ratio) affect noise patterns, we encode this information into our network. 
We create a camera-setting embedding bank with a number (larger than the number of unique camera settings in the dataset) of randomly initialized embeddings, ensuring each setting has a distinct embedding.
Each camera setting's corresponding embedding $\mathbf{z}$ is integrated with the image feature $\mathbf{f}$ using cross-attention. More details are provided in the appendix.

As a result, these embeddings are optimized during training, effectively delivering the corresponding camera setting information to the network and ensuring it produces the appropriate noise pattern. This approach eliminates the need to train separate networks for each camera setting.

\begin{figure*}[t]
  \centering
  \includegraphics[width=1.\linewidth]{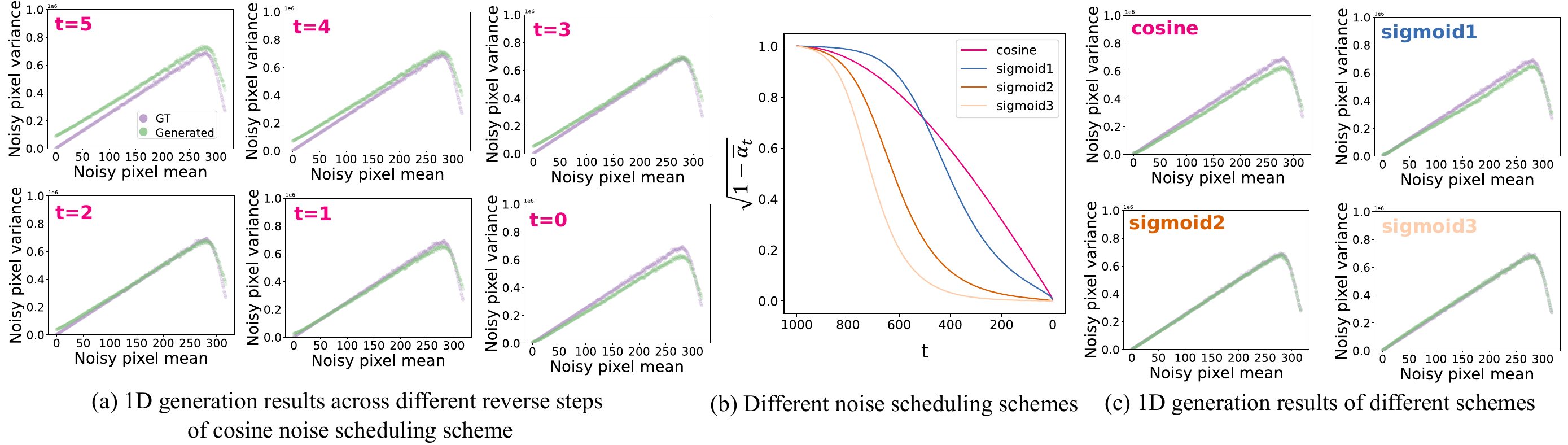}
  \caption{\pami{Impact of noise schedules. (a) Visualization of the mean-variance correspondences of the \textcolor[HTML]{B08DC3}{real} and \textcolor[HTML]{7FBF7C}{generated} Poisson noises across different reverse steps ($t=5\rightarrow 0$) under the cosine noise-scheduling scheme. (b) Different noise scheduling schemes, the y-axis denotes the coefficient $\sqrt{1 - \overline{\alpha}_t}$ of the noise part in Eq.~\ref{eq:x_t} of the appendix. (c) Results of different noise scheduling schemes, where \textcolor[HTML]{D95F02}{\textit{sigmoid2}} and \textcolor[HTML]{FDCDAC}{\textit{sigmoid3}} perform the best in our 1D toy example. Zoom in for a clearer view.}}
  \label{fig:effect_noise_scheduling}
\end{figure*}

\subsection{Variance Preserving Noise Schedule}
\label{sec:mmse_denoising}

Accurately modeling the noise variance is crucial for low-light noise generation as the slightest shift would impair the results of our subsequent denoising task.
In this section, we explore the effects of various diffusion noise schedules and show the importance of selecting an appropriate schedule for generating low-light noise with precise variance.

Diffusion models use a predefined noise schedule to control the amount of Gaussian noise added to the input data (for details, please refer to the appendix). We observed that the commonly used cosine noise schedule~\cite{rombach2022high} reduces the variance of the generated noise.

\noindent \textbf{MMSE Denoising Leads to Reduced Variance.} 
During the reverse diffusion process, the diffusion network is trained to denoise the result from the previous timestep using the mean-squared-error (MSE) loss. 
However, it is known that the output of a denoising network trained with MSE loss is the expectation of all potential clean solutions~\cite{lehtinen2018noise2noise,ledig2017photo,elad2023image}.
Additionally, higher noise levels make the problem more ill-posed, thus resulting in a smoother denoised outcome.

Formally, consider the denoising problem where a measurement \( v \) is related to the signal \( u\sim U \) by \( v = u + \epsilon \), with \( \epsilon \sim \mathcal{N}(0, \sigma_1^2) \) representing the Gaussian noise (added during the forward diffusion process). In our context, the target distribution \( U \) we aim to recover is the low-light noise distribution.
For simplicity, here we assume that $U$ follows a normal distribution $\mathcal{N}(\lambda,\sigma_2^2)$.  
The goal is to learn a denoiser \( f \) (i.e., the diffusion network) to recover \( u \) from \( v \), given knowledge of \( \sigma_1 \).

\begin{proposition}[Proof in the appendix.]
\label{lemma:lemma_1}
When the denoiser is trained using the MSE loss, the optimal solution is given by \( f_{\text{MMSE}}(v) = \mathbb{E}[u | v] \), and the variance of the solution is equal to  $\frac{\sigma_2^4}{\sigma_2^2 + \sigma_1^2}$.
\end{proposition}

As \( \sigma_1^2 \) increases, the variance of the synthesized noise decreases. 
Therefore, to preserve the variance of the signal \( U \), \( \sigma_1^2 \) should be minimized. 
In other words, in the final steps of the reverse process (where the diffusion model focuses more on high-frequency details~\cite{wang2023diffusion}), the diffusion model should be exposed to minimal Gaussian noise; otherwise, the generated output will be overly smoothed. 

\liying{This theory can be easily extended to more complex noise distributions, as the same property holds for Gaussian Mixture Models (GMMs), which can approximate any distribution, given a sufficient number of components. The details can be found in the appendix.
}

Next, we validate this hypothesis through experiments on a 1D toy example by testing different noise schedules.

\noindent \textbf{Performance of Different Noise Schedules.} 
For simplicity, we use a 1D Poisson distribution as a basic approximation of real-world noise.
We train a 1D diffusion model using a 1D UNet to approach this distribution.
The diffusion model employs the commonly used cosine noise schedule, which is plotted in Fig.~\ref{fig:effect_noise_scheduling} (b) in magenta. 

In Fig.~\ref{fig:effect_noise_scheduling} (a), we present the results of ground-truth and generated Poisson noises at different reverse steps (the last 4th, 2nd and final steps are shown for clarity). 
As the variance and mean values of a Poisson distribution are linearly related, we visualize this correspondence.
Note that the visible tails are due to the clipping operation, where pixels saturate. 
We observe that the distribution of the generated results approaches the ground truth as the reverse steps progress~($t\rightarrow0$).
However, after the final reverse step, one can observe that the variance of the generated Poisson noise consistently remains lower than that of the true Poisson distribution.
Based on the explanation provided by Proposition \ref{lemma:lemma_1}, this observation can be attributed to the cosine noise schedule, which introduces excessive Gaussian noise at the beginning of the forward process.
Consequently, at the end of the reverse process, the diffusion network must contend with an excessive amount of noise, leading to overly smooth outputs.

As noise schedule is crucial for accurately mapping the noise variance, we investigate various schedules with smoother slopes at the end of the backward process (with small $t$). 
As depicted in Fig.~\ref{fig:effect_noise_scheduling} (b) and (c), we experiment with four different schemes. 
We observe that the smoother the slope of the curve at the end of the backward process, the better the results are in terms of variance preservation.
Based on these observations, we empirically set the noise schedule to \textit{sigmoid2} for subsequent experiments (\textit{sigmoid3} can also be chosen as it has similar performance).

\liying{In the appendix, we also conduct an experiment on a set of 1D Tukey-Lambda~(TL) distributions to further validate this theory.}

\begin{table*}[t]
    \caption{Denoising performance on the SID and ELD datasets in terms of PSNR and SSIM across various exposure ratios. Noise synthesis methods are divided into two groups: \textit{W/ Calibration}, which use dark shading data while training the denoising network; and \textit{W/o Calibration}, which do not use dark shading.
    Best results are highlighted in \best{bold}, and second-best results are \second{underlined}. Our \textit{NoiseDiff} and \textit{NoiseDiff*} outperform other noise synthesis methods in most cases under the same evaluation setting.
    }
  \centering
  \resizebox{1.0\textwidth}{!}{
  \begin{tabular}{cc>{\columncolor{gray!20}}c >{\columncolor{lightcyan}}c>{\columncolor{lightcyan}}c>{\columncolor{lightcyan}}c>{\columncolor{lightcyan}}c >{\columncolor{lightcyan}}c >{\columncolor{gray!20}}c >{\columncolor{gray!20}}c >{\columncolor{gray!20}}c}
  
  \toprule
  \multirow{2}{*}{\textbf{Dataset}}  & \multirow{2}{*}{\textbf{Ratio}} &  \multicolumn{1}{c}{\textit{\textbf{Real-noise-based}}} &  \multicolumn{5}{c}{\textbf{W/o Calibration}} & \multicolumn{3}{c}{\textbf{W/ Calibration}} \\ \cmidrule(r){3-3} \cmidrule(r){4-8} \cmidrule(r){9-11}
   &  & \cellcolor{white} \textit{\textbf{Real data}} & \cellcolor{white} \textbf{Poisson-Gaussian} & \cellcolor{white} \textbf{ELD}~\cite{wei2021physics} & \cellcolor{white} \textbf{\pami{Starlight}}~\cite{monakhova2022dancing} & \cellcolor{white} \textbf{NoiseFlow}~\cite{abdelhamed2019noise} & \cellcolor{white} \textbf{NoiseDiff} & \cellcolor{white} \textbf{PMN}~\cite{feng2023learnability} & \cellcolor{white} \textbf{LRD}~\cite{zhang2023towards} & \cellcolor{white} \textbf{NoiseDiff*} \\ \midrule
   
    \multirow{3}{*}{\textbf{SID}} 
    & $\times$100 & \textit{42.95 / 0.958} & 41.05 / 0.936 & \second{41.95} / \second{0.953} & 40.47 / 0.926 & 40.20 / 0.925 & \best{43.30} / \best{0.958} & \second{43.47} / \best{0.961} & 43.16 / 0.958 & \best{43.92} / \best{0.961} \\ 
    & $\times$250 & \textit{40.27 / 0.943} & 36.63 / 0.885 & \second{39.44} / \second{0.931} & 36.25 / 0.858 & 36.15 / 0.870 & \best{40.53} / \best{0.944} & \second{41.04} / \best{0.947} & 40.69 / 0.941 & \best{41.28} / \second{0.946} \\ 
    & $\times$300 & \textit{37.32 / 0.928} & 33.34 / 0.811 & \second{36.36} / \second{0.911} & 32.99 / 0.780 & 33.27 / 0.803 & \best{37.68} / \best{0.928} & \second{37.87} / \best{0.934} & 37.48 / 0.919 & \best{37.94} / \second{0.930} \\ 
    
    \cdashlinelr{1-11}
    \multirow{2}{*}{\textbf{ELD}} 
    & $\times$100 & \textit{45.52 / 0.977} & 44.28 / 0.936 & \second{45.45} / \best{0.975} & 43.80 / 0.936 & 43.31 / 0.941 & \best{45.79} / \second{0.972} & \best{46.99} / \best{0.984} & 46.16 / 0.983 & \second{46.95} / \second{0.978} \\ 
    & $\times$200 & \textit{41.70 / 0.912} & 41.16 / 0.885 & \best{43.43} / \best{0.954} & 40.86 / 0.884 & 40.26 / 0.885 & \second{42.25} / \second{0.924} & \second{44.85} / \second{0.969} & 43.91 / 0.968 & \best{45.11} / \best{0.971} \\
   
   \bottomrule 
  \end{tabular}}
\label{tab:compare_sota}
\end{table*}

\section{Experimental Results}\label{sec:experiments}

\subsection{Evaluation Protocols}
\label{sec:metrics}

\noindent \textbf{Image Denoising Performance.}
A direct approach to assess the quality of the noisy images generated by our model is to train a denoising network with them. 
We can then evaluate its performance on standard low-light denoising test sets.
We compute the Peak Signal-to-Noise Ratio (PSNR) and Structural SIMilarity (SSIM) between the denoised and ground-truth images in the RAW domain.

Since PSNR and SSIM metrics are sensitive to overall image illumination, even minor shifts can lead to significant variations in these values. 
We apply the illumination correction method from the ELD paper~\cite{wei2021physics} \footnote{See this part of the \href{https://github.com/Vandermode/ELD/blob/78156f277fe8b3b10b8b16596afea6854d59e666/models/ELD_model.py\#L138-L169}{code}.} to the denoised images.
When comparing our method with other state-of-the-art methods, we apply this illumination correction to all methods before evaluating PSNR and SSIM.

Note that the original ELD paper crops the image centers to a size of $512 \times 512$ for PSNR/SSIM computation, whereas we compute them on the entire image at a resolution of $4240\times2832$.
In the following results, we apply the same measurement setup for all methods.

\noindent \textbf{Noise Statistics.}
We also analyze the pixel-wise marginal distributions between the generated and real noise images in the test set, similar to the approach presented in the Noise Flow paper~\cite{abdelhamed2019noise}.
This evaluation prioritizes the model’s ability to capture the fundamental characteristics of the noise distribution. 
Specifically, we compute the histograms of real-world and generated noise and report the discrete Kullback-Leibler (KL) divergence between these histograms.

We further analyze the characteristics of the generated noise by evaluating its variance and mean values.
For each possible clean pixel value, we gather the corresponding noise pixels and compute their mean and variance. The specific process is given in the appendix.

\subsection{Implementation Details}

We generate noise images in a patch-wise manner.
Since our diffusion network is trained on $512 \times 512$ patches, the generated noise patches are also $512 \times 512$.
We use DDPM sampling~\cite{ho2020denoising} for inference.

We train our diffusion model using the SID Sony RAW image training set. 
For evaluation, in line with the approach used in the ELD paper~\cite{wei2021physics}, we evaluate our denoising network on a combination of the SID test and validation sets (referred to as the SID test set hereafter), using the same evaluation images as in the ELD paper. 
We also evaluate our model on the ELD Sony dataset.

\liying{To assess the quality of the synthetic data, we train denoising networks using the $L1$ loss between the network's output and the reference clean image, then evaluate their denoising performance. 
For a fair comparison, we use the same U-Net architecture employed in \cite{chen2018learning}.}

Further implementation details are provided in the appendix.

\begin{figure*}[t]
  \centering
  \includegraphics[width=1.0\linewidth]{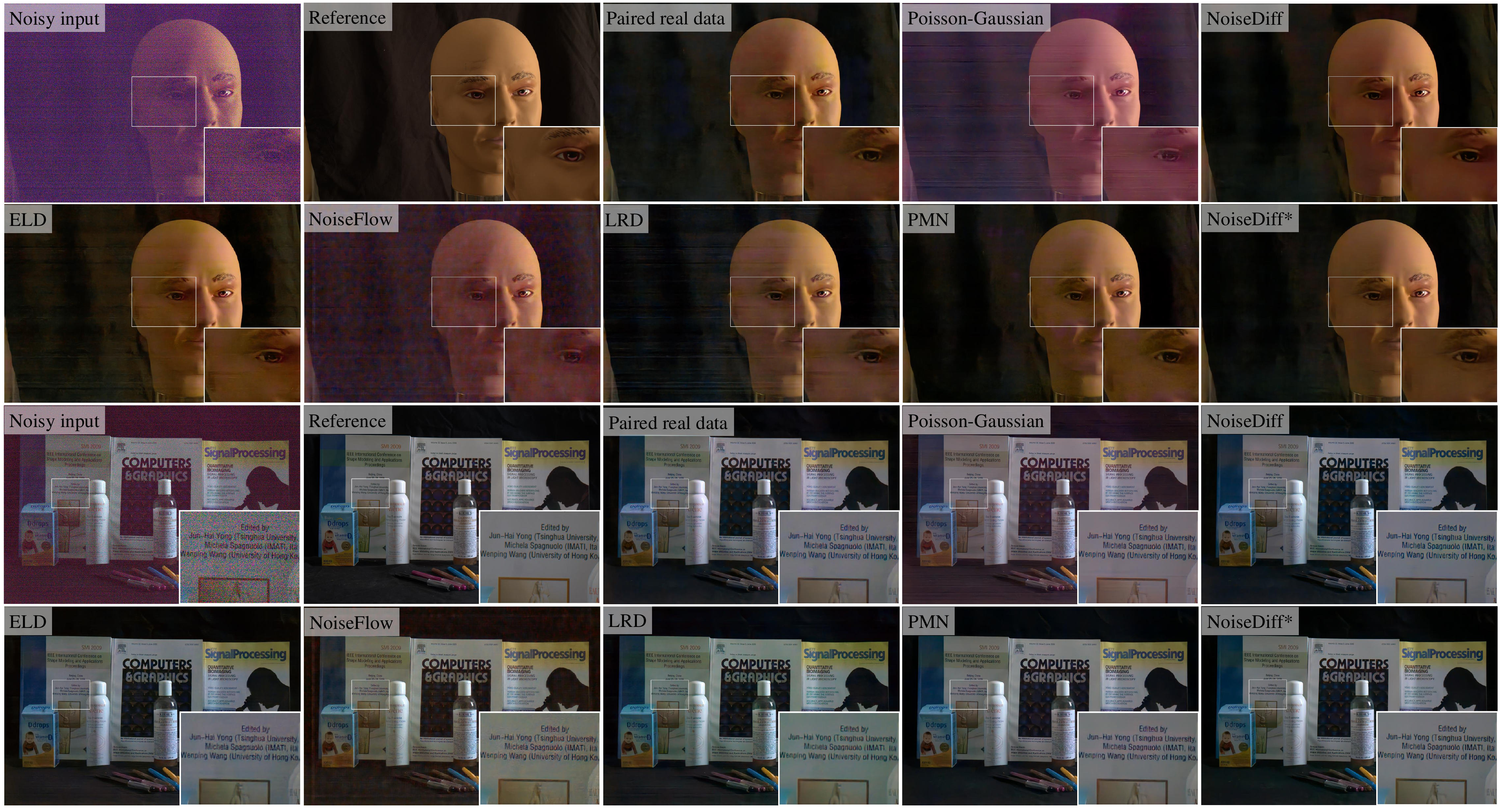}
  \caption{Visual comparison of denoising results from networks trained on data synthesized using different methods, alongside comparisons with a model trained on real clean-noisy pairs from the SID training set. \liying{The first sample is from the SID test set, while the second is from the ELD test set. 
  In the first example, our method produces cleaner results, while other methods leave some residual noise. The second example demonstrates that our method recovers text details more clearly and with fewer artifacts.
  Best viewed when zoomed in.} 
  }
  \label{fig:comparison_sid_eld}
\end{figure*}

\pami{

\begin{table}[t]
  \caption{\pami{KL divergence comparisons of generated noise on the SID dataset of different data synthesis methods.}}
  \centering
    \resizebox{0.95\columnwidth}{!}{
    \begin{tabular}{c|cccc}
        \hline
        & Poisson-Gaussian & NoiseFlow & PMN & NoiseDiff \\
        \hline
        $\times$100 & 0.094 & 0.415 & 0.139 & \textbf{0.025} \\
        $\times$250 & 0.070 & 0.193 & 0.196 & \textbf{0.031} \\
        $\times$300 & 0.187 & 0.155 & 0.407 & \textbf{0.057} \\
        \hline
    \end{tabular}}
    \label{tab_reb:kld}
\end{table}

\subsection{Noise statistic Comparison}

To evaluate the quality of noise generated by different methods, we compare the KL divergence between the real noise distribution and the generated noise distributions of different data synthesis methods. The results are presented in Table~\ref{tab_reb:kld}, excluding methods without released noise generation code. The results show that our method achieves the lowest KLD.

}

\subsection{Image Denoising Performance Comparison}
\label{sec:compare}
As stated in Sec.~\ref{sec:metrics}, we compare the performance of denoising networks trained with data from different synthesis methods.
\liying{
We evaluate three physics-based methods (the Poisson-Gaussian model, ELD~\cite{wei2021physics} and PMN~\cite{feng2023learnability}), and three recent learning-based methods (\pami{Starlight\cite{monakhova2022dancing},   } LRD~\cite{zhang2023towards} and NoiseFlow~\cite{abdelhamed2019noise})\footnote{For ELD, PMN and LRD, we use pre-trained models available on the authors’ official websites. For NoiseFlow, we retrain it on the SID dataset. \pami{The results of Starlight are taken from the LLD paper\cite{cao2023physics}}.}.
We also compare our method with a denoising network trained with real data pairs from the SID Sony training set. 

Among these methods, PMN and LRD apply dark shading correction (DSC) for noise calibration, where a precomputed dark shading frame—obtained by averaging multiple dark frames, as described in Sec.~\ref{sec:rel_works}—is subtracted from the input noisy images during both denoising training and inference. In contrast, the Poisson-Gaussian model, ELD, and NoiseFlow do not incorporate this correction.

For a fair comparison, we train two versions of our denoising network. The first, \textit{NoiseDiff}, is trained directly on our synthetic data without DSC. The second, \textit{NoiseDiff*}, is trained on our synthetic data processed with DSC by subtracting the dark shading.

Additionally, \textit{NoiseDiff*} employs shot noise augmentation (SNA) proposed in PMN.
This method relies on the linear behavior of shot noise, by artificially scaling the ground-truth image and adding Poisson noise accordingly to the noisy images. Specifically, the clean image $I$ is first scaled by a random factor $\delta$, resulting in a \textit{clean signal increment} $\Delta = (1-\delta)I$. A \textit{noisy signal increment} is then sampled from a Poisson distribution $\mathcal{P}(\Delta)$ and added to the noisy image, in line with the image formation process described in Sec.~\ref{sec:rel_works}.
In practice, SNA enhances the diversity of shot noise in our synthetic data. In LRD, shot noise is sampled from a Poisson distribution on the fly during training, making this process equivalent to SNA. This equivalence justifies the comparison with our method augmented with SNA.
For more details about DSC and SNA, please refer to the appendix.


It is also worth noting that, in addition to augmented data, PMN also uses real data pairs from the SID Sony training set during training of the denoising network. In contrast, all other data synthetic methods, including our \textit{NoiseDiff} and \textit{NoiseDiff*}, use only synthetic data.
}

\noindent \textbf{Numerical Results.}
\liying{Table~\ref{tab:compare_sota} presents quantitative comparisons between different methods. The middle part compares \textit{NoiseDiff} without calibration to other calibration-free methods, while the right part evaluates \textit{NoiseDiff*} against methods that incorporate dark shading data for calibration.}

\liying{
As shown in Table~\ref{tab:compare_sota}, our \textit{NoiseDiff} achieves the best results on the SID dataset among calibration-free data synthesis methods and outperforms models trained with real data pairs in most cases. Our method also demonstrates competitive PSNR/SSIM performance on the ELD dataset.

Dark shading calibration eases the training of the denoising network, as it centers noise on zero, by removing the inhomogeneous bias in dark frames. Therefore, DSC calibrated denoising networks tend to perform better than previous ones.
Among them, our \textit{NoiseDiff*} achieves the best performance in most cases. For example, on SID $\times 100$, it surpasses PMN, the second-best approach, by 0.45 dB in PSNR.
}

\noindent \textbf{Qualitative Results.}
\liying{
We compare denoising performance on two examples from the SID and ELD test sets in Fig.~\ref{fig:comparison_sid_eld}. In the zoomed-in area of the SID example, our method produces cleaner results, while other methods leave some residual noise. The second example from the ELD test set demonstrates that our method recovers text details more clearly and with fewer artifacts. Additional examples in the appendix show that, while other calibration-free noise synthesis methods struggle with low-frequency noise appearing at the bottom of the image, our method effectively addresses this issue even without calibration data.
}

\begin{figure*}[t]
  \centering
  \includegraphics[width=1.\linewidth]{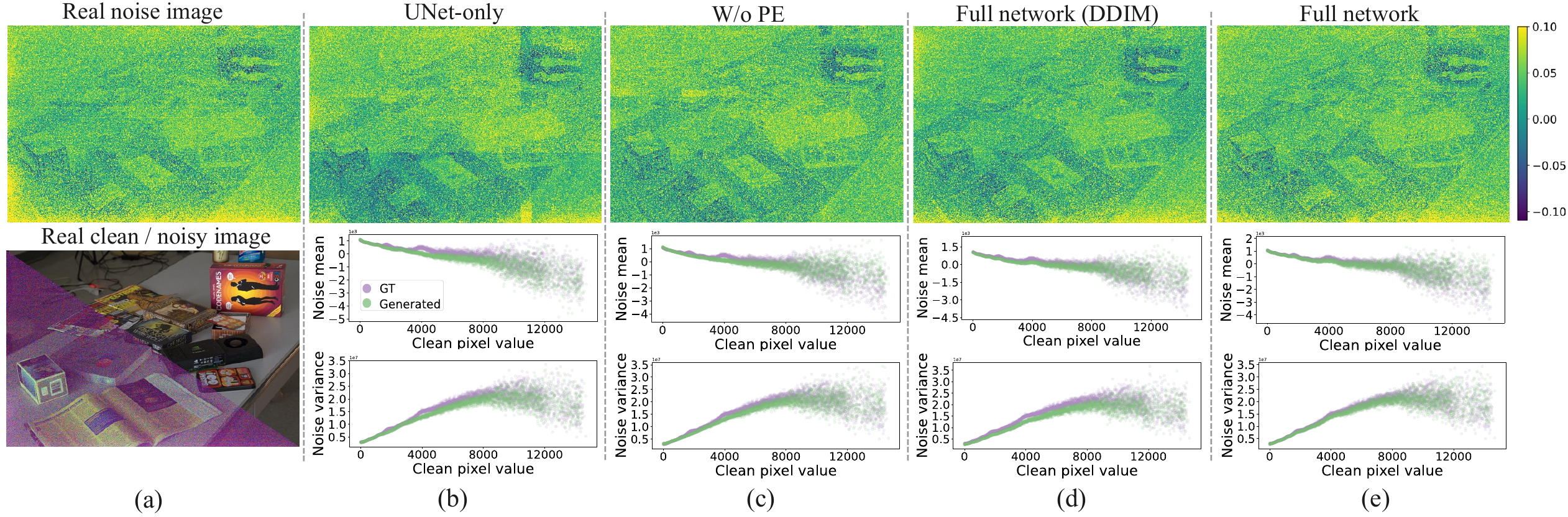}
  \caption{\pami{Ablation study on network architectures and diffusion sampling methods.
  (a) Second row: a real clean-noisy pair from the SID Sony test set (ISO 10,000, ratio 250); first row: the corresponding real noise image.
  (b) Results from the UNet-only architecture. First row: generated noise image (by piecing $512\times512$ patches together); second row: comparison of the mean and variance of the \textcolor[HTML]{7FBF7C}{generated noise} and the \textcolor[HTML]{B08DC3}{real noise}.
  (c) Results from the network without positional encoding.
  (d) Results from the full network using DDIM sampling.
  (e) Results from our full network using DDPM sampling.
  The UNet-only architecture produces noise with inaccurate mean and variance. Without positional encoding, the network fails to model the low-frequency noise at the image bottom. DDIM sampling leads to reduced noise variance. Our full model generates the most accurate noise pattern.}
  }
  \label{fig:ablation_noise_visualization}
\end{figure*}

\subsection{Discussions}
\label{sec:discussions}

In this section, we assess the importance of various components in our diffusion model by analyzing the characteristics of the generated noise and the denoising performance of networks trained on data generated by ablated models. \liying{In the following ablation experiments, we exclude dark shading correction, shot noise augmentation, and illumination correction when assessing denoising performance. This ensures that the results reflect the direct network output, making it easier to isolate and evaluate the effects of our main contributions.
}


\begin{table}[t]
  \caption{Ablation study on the influence of training-data diversity for denoising. We show the PSNR/SSIM results on the SID test set.}
  \centering
  \begin{tabular}{c||c|c}
      \hline
      Ratio & NoiseDiff (Limited data) & NoiseDiff \\
      \hline
      $\times 100$ & 42.42 / 0.956 & \best{42.55} / \best{0.957}  \\
      $\times 250$ & \best{39.80} / 0.941 & 39.67 / \best{0.943}  \\
      $\times 300$ & 36.87 / 0.924 & \best{37.24} / \best{0.927} \\
      \hline
  \end{tabular}
  \label{tab:aba_train_number}
\end{table}

\noindent \textbf{Influence of Training-Data Diversity.}
The number of scenes per camera setting in the SID training set is uneven, with some settings having more than ten scenes and some only one (e.g., 28 scenes for ISO 250 and 1 for ISO 2500). To evaluate the influence of training-data diversity, we train a denoising network on synthetic data using the same number of scenes per camera setting as in the SID training set. The result is reported as \textit{NoiseDiff (Limited data)}.

To leverage the full potential of our diffusion model, we generate more synthetic data by randomly sampling scenes from the SID dataset, regardless of their camera settings. This sampling strategy improves image denoising performance, as shown in Table \ref{tab:aba_train_number}.
Visual comparisons in the appendix further support that it improves the results for under-represented camera settings.
Thus, our method offers a promising approach for improving denoising performance when real-world data is scarce.

\noindent \textbf{Sampling Method.}
The generation speed of our method is constrained by the many steps required by DDPM sampling. \pami{We tested our model on a server with dual AMD EPYC 7543 32-core CPUs (128 threads total) and an NVIDIA A100-SXM4-80GB GPU. The average time for generating a $512\times512$ patch using DDPM (1000 steps) is $43.6~s$, excluding IO time.}

Therefore, in the following ablation studies, we use DDIM sampling~\cite{song2020denoising} as an alternative, which uses a non-Markovian framework to achieve faster generation with fewer steps. 
We first validate that DDIM could achieve satisfactory performance with a minor drop compared to DDPM. 
We train a denoising network with synthetic data generated using 100-step DDIM sampling. As shown in Fig.~\ref{fig:ablation_noise_visualization} (d), DDIM generates noise resembling real noise, with slight variance inaccuracy.
Table \ref{tab:aba_arch} (C) and (D) show that while DDIM introduces a performance trade-off, the decrease is acceptable.

\begin{figure*}[htp]
  \centering
  \includegraphics[width=1\linewidth]{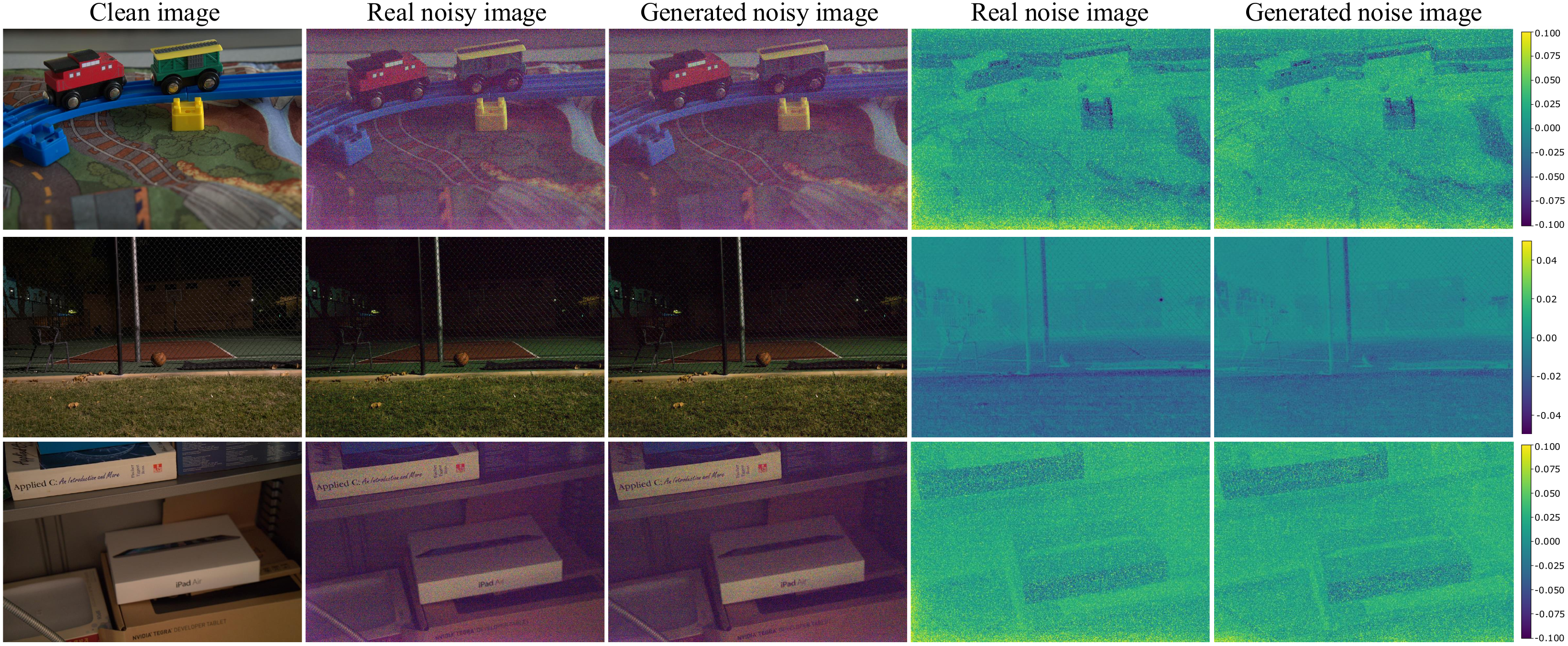}
  \caption{Comparison between real and synthetic data. The samples in the three rows are from the SID Sony test set (not used for diffusion training) under the following settings: ISO 6,400 with a ratio of 300; ISO 200 with a ratio of 250; and ISO 12800 with a ratio of 100. Different noise values are represented by varying colors, with the color bar displayed on the right. Our method generates realistic noise that closely resembles real noise.
  }
  \label{fig:noise_visualization}
\end{figure*}

\begin{table}[t]
    \caption{Ablation study on network architectures and diffusion sampling methods.
    We present the PSNR/SSIM results of denoising networks trained with different synthetic data sources generated by various methods on the SID test set. We also report the KL divergence (in \textit{italic}) between the noise distributions produced by these methods and the real noise distribution.}
    \centering
    \resizebox{1.\columnwidth}{!}{
    \begin{tabular}{c||cc|c:c}
        &(A) & (B) & (C) & (D)\\
        \hline
        Exposure&   \textit{UNet-only}  & \textit{W/o PE}  & \textit{Full network } & \textit{Full network} \\
        Ratio&   \textit{(DDIM)} & \textit{(DDIM)} & \textit{(DDIM)} & \textit{(DDPM)} \\
        \midrule
        \multirow{2}{*}{$\times 100$} &  41.45 / 0.953 & 41.64 / 0.948 & 42.03 / 0.953 & \textbf{42.42 / 0.956} \\
                                   & \textit{0.645} & \textit{0.622} & \textit{0.622} & \textbf{\textit{0.568}} \\
        \hline
        \multirow{2}{*}{$\times 250$}  & 39.22 / 0.936 & 38.94 / 0.925 & 39.44 / 0.937 & \textbf{39.80 / 0.941} \\
                                    & \textit{0.016} & \textit{0.011} & \textit{0.008} & \textit{\textbf{0.004}} \\
        \hline
        \multirow{2}{*}{$\times 300$}  & 36.59 / 0.915 & 35.99 / 0.898  & 36.67 / 0.916 & \textbf{36.87 / 0.924} \\
                                       & \textit{0.028} & \textit{0.022} & \textit{0.019} & \textit{\textbf{0.018}} \\
        \hline
    \end{tabular}}
    \label{tab:aba_arch}
\end{table}

\noindent \textbf{Effects of Two-Branch Network Architecture.}
\label{exp:two_branch}
We investigate the effects of our proposed two-branch network architecture by conducting an ablation study, which adopts a sole UNet architecture for the diffusion model. 
This UNet-only architecture takes the clean image $\mathbf{x}$ and the output of the last step $\Tilde{\mathbf{n}}_{t}$ as inputs; it then outputs the result of the current step $\Tilde{\mathbf{n}}_{t-1}$. 

To assess performance differences, 
\liying{we train this UNet-only model from scratch using the same training setup as the full network.}
We then generate synthetic data from clean images in the SID training set using both the full network and the UNet-only network, employing DDIM sampling for efficiency. We compare the performance of two denoising networks trained on these respective synthetic datasets, as shown in Table~\ref{tab:aba_arch}. 
Comparing (A) and (C) shows that the full network achieves higher PSNR/SSIM. Additionally, the full model's noise has a smaller KL divergence from real noise than the UNet-only model.  
\pami{Note that the KLD values for our full model in Table~\ref{tab:aba_arch} differ slightly from those reported in Table~\ref{tab_reb:kld}. This is because, for efficient ablation studies, we sample a subset of the data instead of using the full dataset. In contrast, Table~\ref{tab_reb:kld} reports the KLD calculated over the entire dataset. }

We also condition both architectures on the same clean images from the SID test set and visualize the generated results. Examples are shown in Fig.~\ref{fig:ablation_noise_visualization} (b) and (e). 
Note that our diffusion network generates patches of $512\times 512$ size. 
To visualize the full image, we piece patches together. 
The noise patches produced by the UNet-only architecture exhibit a checkerboard pattern; this indicates inconsistent mean values among patches. This phenomenon suggests that the UNet-only architecture generates noise images with unstable mean values.

\noindent \textbf{Effects of Positional Encoding.}   We further investigate the effects of positional encoding. 
We conduct the following experiment to assess its effects: During inference, we modify the generation process to $\Tilde{\mathbf{n}}_{t} = f_{\theta} (\Tilde{\mathbf{n}}_{t-1}, t, \mathbf{x}, \mathbf{0}, \mathbf{z})$, where $\mathbf{0}$ denotes a black image filled with zeros, signifying that the diffusion model is conditioned on a zero map instead of on the exact pixel coordinates. 
We compare KL divergence of noise generated by models with and without positional encoding. 
Table~\ref{tab:aba_arch} (B) and (C) show that the denoising network trained with data generated with positional encoding yields higher PSNR and SSIM, and the KL divergence increases when positional encoding is not used.
The generated results are depicted in Fig.~\ref{fig:ablation_noise_visualization} (c). By comparing it with the full results in Fig.~\ref{fig:ablation_noise_visualization} (e), it can be observed that the generated noises do not contain the low-frequency noise in the bottom; this indicates the effectiveness of positional encoding in modeling the position-related noises.

\noindent \textbf{Noise Visualization.} 
In Fig.~\ref{fig:noise_visualization}, we visualize noise images generated by our model. By comparing real and generated noise, we demonstrate that our diffusion model accurately captures realistic noise patterns. In the first sample, the real noise image exhibits low-frequency noise at the bottom.
Our model successfully reproduces this, as the generated noise also shows low-frequency noise at the bottom. Failure to capture this pattern would result in a purplish offset in the final denoised images. \liying{As shown in some examples in the appendix}, our denoising network effectively handles low-frequency noise, unlike other methods, which produce purplish artifacts at the bottom.

\pami{One can observe that in our generated noise images, the banding noise is visually less prominent compared to the real data. In real noise, the banding pattern appears as long horizontal lines, while in our synthetic noise images, the bands are shorter and less continuous. This may be attributed to the limited receptive field of our network and the inherent randomness of the diffusion sampling process.
Nonetheless, as demonstrated by our qualitative denoising results (e.g., Fig.~\ref{fig:comparison_sid_eld}), our denoiser effectively removes banding noise, while competing methods such as LRD and ELD leave residual artifacts. This indicates that the shorter synthetic banding pattern in our data is sufficient for training a denoiser capable of handling the more pronounced banding pattern observed in real data.
}

\noindent \textbf{Noise Decomposition.} Our framework enables noise decomposition, allowing for further analysis of generated-noise characteristics by separating it into signal-dependent and signal-independent components. 
Our findings suggest that this method can implicitly produce dark frames, and it points toward a potential approach for efficient data generation in future work. 
See the appendix for details.

\section{Limitations}
Despite the promising results, our method has some limitations. \pami{As noted in the discussion, the banding noise in our generated noise images is visually less prominent compared to the real data.}
The generation speed is also relatively slow due to the many steps required by the DDPM sampling process. Although DDIM sampling offers some acceleration, it does not map the noise variance as accurately. Consequently, faster and more accurate approaches remain an area for further exploration.
Additionally, our method still relies on a certain number of paired clean-noisy images for data synthesis training. Investigating unsupervised or self-supervised approaches could be a promising direction for future work.

\section{Conclusion}
Diffusion models have been widely applied for generating realistic images, but their use for generating real-world noise remains unexplored. 
Nonetheless, we find that a direct application of diffusion models is insufficient for realistic noise synthesis. 
Our tailored diffusion framework—featuring a two-branch architecture, pixel positional encoding, and carefully chosen noise schedules—enables the generation of synthetic noise that closely matches real-world noise, outperforming existing noise synthesis methods.
Moreover, our analysis of the generated data provides valuable insights into leveraging diffusion models for real-world noise synthesis.





\ifpeerreview \else

\begin{IEEEbiography}[{\includegraphics[width=1in,height=1.25in,clip,keepaspectratio]{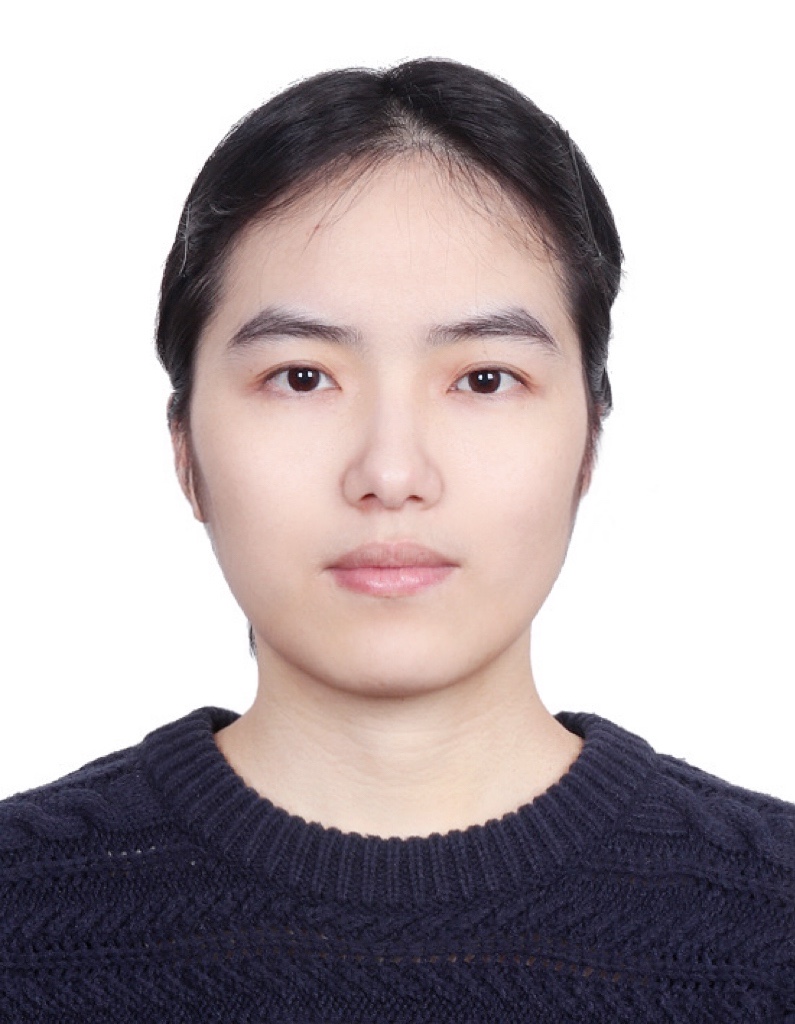}}]{Liying Lu} received her B.E. degree from Huazhong University of Science and Technology (HUST) and the M.Phil. degree from the Chinese University of Hong Kong (CUHK). She is currently pursuing the Ph.D. degree at the École Polytechnique Fédérale de Lausanne (EPFL) in the Image and Visual Representation Lab.
Her research interests include computer vision, computational photography, and computational imaging. 
\end{IEEEbiography}

\begin{IEEEbiography}[{\includegraphics[width=1in,height=1.25in,clip,keepaspectratio]{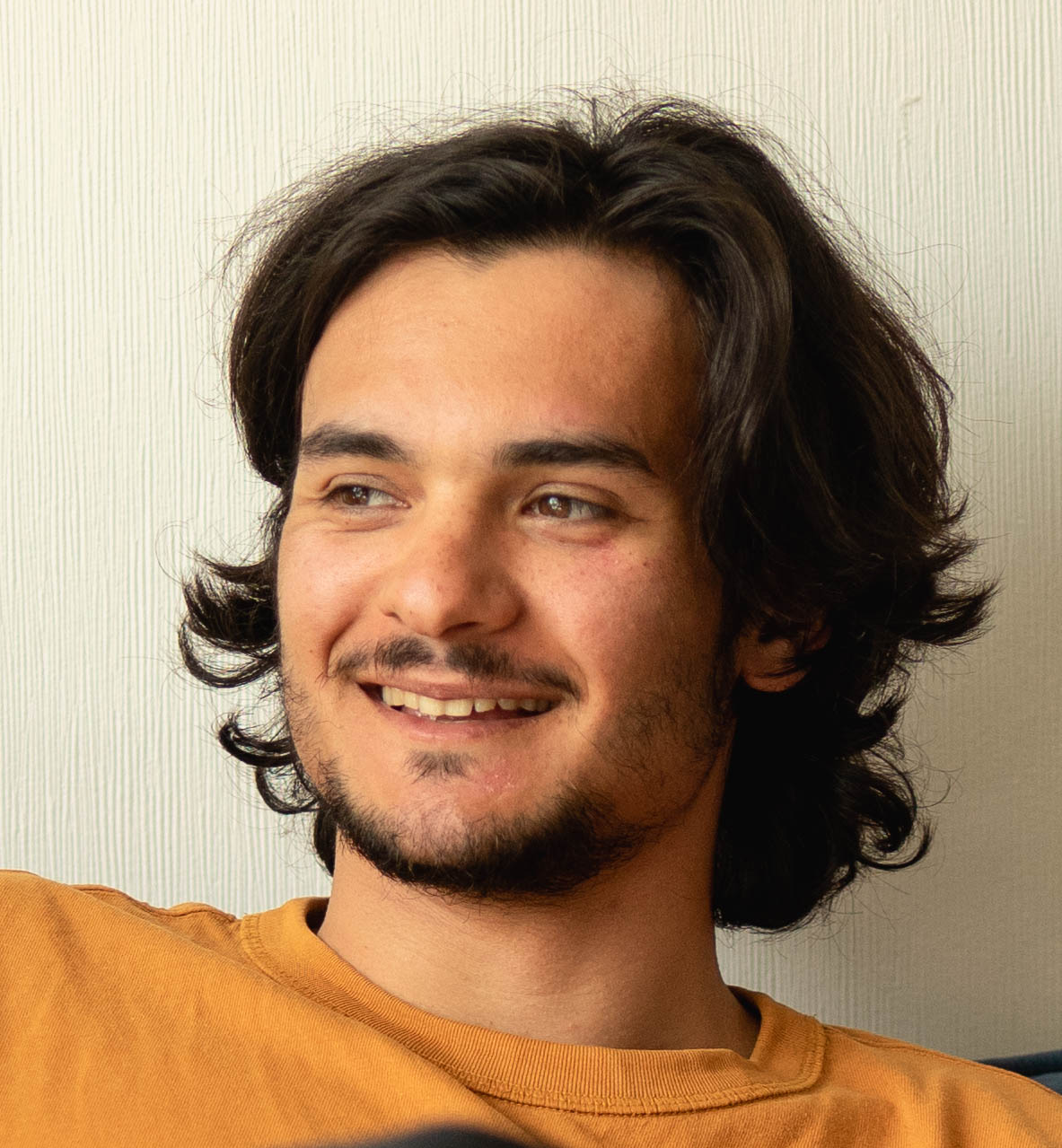}}]{Raphaël Achddou} received his Ph.D. degree in Computer Science from Telecom Paris, Institut Polytechnique de Paris, in 2023. He is currently a Post-Doctoral researcher at the Image and Visual Representation Lab at the École Polytechnique Fédérale de Lausanne (EPFL). His research interests include image restoration, computational photography, statistical modeling of natural images, and computer vision. He served as the local arrangement chair for ICCP 2024.
\end{IEEEbiography}

\begin{IEEEbiography}[{\includegraphics[width=1in,height=1.25in,clip,keepaspectratio]{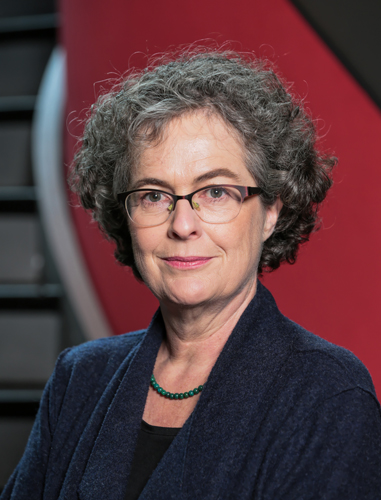}}]{Sabine Süsstrunk} leads the Image and Visual Representation Lab in the School of Computer and Communication Sciences at École Polytechnique Fédérale de Lausanne (EPFL). Her main research areas are in computational photography, computational imaging, generative AI, and computational image quality and aesthetics. 
She is President of the Swiss Science Council SSC, Founding Member and Member of the Board (President 2014-2018) of the EPFL-WISH (Women in Science and Humanities) Foundation, Member of the Board of the SRG SSR (Swiss Radio and Television Corporation), and Co-Founder and Member of the Board of Largo Films SA. She received the IS\&T/SPIE 2013 Electronic Imaging Scientist of the Year Award for her contributions to color imaging, computational photography, and image quality. Sabine is a Fellow of AIAA, ELLIS, IEEE and IS\&T and member of SATW. She was general chair of ICCP 2024.
\end{IEEEbiography}




\fi

\twocolumn[{%
\renewcommand\twocolumn[1][]{#1}%
\maketitle
\section*{\centering \LARGE Appendix: Dark Noise Diffusion \\ Noise Synthesis for Low-Light Image Denoising}

\vspace{15pt}
This appendix is organized as follows:

\begin{itemize}
    \item Preliminaries: Diffusion Models $\rightarrow$ Sec.~\ref{supp:diffusion}
    \item Proof for Variance Preserving Noise Schedule $\rightarrow$ Sec.~\ref{supp:proof}
    \item Implementation Details $\rightarrow$ Sec.~\ref{supp:implementation}
    \begin{itemize}
        \item Network Architecture $\rightarrow$ Sec.~\ref{supp:architecture}
        \item Datasets $\rightarrow$ Sec.~\ref{supp:datasets}
        \item Training Configurations $\rightarrow$ Sec.~\ref{supp:training}
        \item Clean-Noisy Pair Generation $\rightarrow$ Sec.~\ref{supp:pair_gen}
    \end{itemize}
    \item Evaluation Protocol: Noise Mean and Variance $\rightarrow$ Sec.~\ref{sec:supp_evaluation_protocol}
    \item More Discussions $\rightarrow$ Sec.~\ref{supp:more_dis}
    \begin{itemize}
        \item Influence of Training-Data Diversity $\rightarrow$ Sec.~\ref{supp:data_diversity}
        \item Sampling Method $\rightarrow$ Sec.~\ref{supp:sampling}
        \item Noise Decomposition $\rightarrow$ Sec.~\ref{supp:noise_decomp}
        \item Noise Histogram $\rightarrow$ Sec.~\ref{supp:noise_hist}
        \item Influenc of Dark Shading Correction and Shot Noise Augmentation $\rightarrow$ Sec.~\ref{supp:dsc_sna}
    \end{itemize}
    \item More Visual Comparisons $\rightarrow$ Sec.~\ref{supp:more_visual}
\end{itemize}

For optimal visual comparisons, we recommend viewing the results on a screen. We will release our code upon acceptance.

\vspace{20pt}
}]

\renewcommand{\thefigure}{S\arabic{figure}}
\setcounter{figure}{0} 
\renewcommand{\thetable}{S\arabic{table}}
\setcounter{table}{0} 

\section{Preliminaries: Diffusion Models}\label{sec:preliminaries}
\label{supp:diffusion}

In this section, we review diffusion models~\cite{ho2020denoising}. Diffusion models involve a forward process and a reverse process. The \textit{forward process}, also called \textit{diffusion process}, formulates the posterior distribution $q(x_{1:T} | x_0)$ as a Markov chain that gradually adds Gaussian noises to the input data $x_0$, according to a noise schedule $\beta_1,...,\beta_T$:
\begin{align}
q(\mathbf{x}_{1:T} | \mathbf{x}_0) &= \prod_{t=1}^{T} q(\mathbf{x}_t | \mathbf{x}_{t-1}) \;, \\
q(\mathbf{x}_t | \mathbf{x}_{t-1}) &= \mathcal{N}(\mathbf{x}_t; \sqrt{1 - \beta_t}\mathbf{x}_{t-1}, \beta_t \mathbf{I}) \label{eq:posterior} \;,
\end{align}
where \( \mathbf{x}_t \) is the noised data at timestep \( t \), and \( \mathcal{N} \) denotes the Gaussian distribution. An important property of the forward process is that we can sample $\mathbf{x}_t$ at an arbitrary timestep $t$, directly from $\mathbf{x}_0$ by
\begin{align}
\label{eq:x_t}
    \mathbf{x}_t = \sqrt{\overline{\alpha}_t}\mathbf{x}_0 + \sqrt{(1 - \overline{\alpha}_t)} \mathbf{\epsilon}_t \;,
\end{align}
where $\mathbf{\epsilon}_t \sim \mathcal{N}(0, \mathbf{I})$, $\alpha_t = 1 - \beta_t$, and $\overline{\alpha}_t = \prod_{s=1}^t \alpha_i$.

The \textit{reverse process} estimates the joint distribution $p_{\theta}(\mathbf{x}_{0:T})$, starting from $\mathbf{x}_T \sim \mathcal{N}(0, \mathbf{I})$, it can be formulated as follows:
\begin{align}
    p_{\theta}(\mathbf{x}_{0:T}) &= p_{\theta}(\mathbf{x}_{T}) \prod_{t=1}^{T} p_{\theta}(\mathbf{x}_{t-1} | \mathbf{x}_t) \;, \\
    where~p_{\theta}(\mathbf{x}_{t-1} | \mathbf{x}_t) &= \mathcal{N}(\mathbf{x}_{t-1};\mathbf{\mu}_{\theta}(\mathbf{x}_t, t), \Sigma_{\theta} \mathbf{I}) \;,
\end{align}
where $\Sigma_{\theta} = \frac{1 - \overline{\alpha}_{t-1}}{1 - \overline{\alpha}_t} \beta_t$, and according to Ho~\etal~\cite{ho2020denoising}, $\mathbf{\mu}_{\theta}(\mathbf{x}_t, t)$ can be obtained by estimating the noise $\mathbf{\epsilon}_t$ added to $\mathbf{x}_t$, through a neural network $\mathbf{\theta}$:
\begin{align}
    \mathbf{\mu}_{\theta}(\mathbf{x}_t, t) = \frac{1}{\sqrt{\alpha_t}} (\mathbf{x}_t - \frac{\beta_t}{\sqrt{1 - \overline{\alpha}_t}} \mathbf{\epsilon}_{\theta}(\mathbf{x}_t, t)) \;.
\end{align}
The network parameters $\theta$ are optimized by minimizing the following MSE loss:
\[
\mathcal{L} = \mathbb{E}_q \left [ || \mathbf{\epsilon}_t -\epsilon_{\theta} (\mathbf{x}_t, t) ||^2_2 \right ] \;. 
\]

\section{More Details about Variance Preserving Noise Schedule}

In this section, we first provide the proof of Proposition~1 discussed in the main paper. We then present an additional experiment on 1D Tukey-Lambda distributions to further validate the importance of noise schedules.

\subsection{Proof for Proposition 1}
\label{supp:proof}


\begin{proof}
Given a signal \( u \sim \mathcal{N}(\lambda, \sigma_2^2) \) and its measurement \( v = u + \epsilon \), where \( \epsilon \sim \mathcal{N}(0, \sigma_1^2) \), \pami{
the conditional distribution \( p(U | V) \) can be derived using properties of Gaussian distributions and Bayesian inference~\cite{bishop2006pattern}. It is given by}

\begin{align}
    p(U|V = v) \sim \mathcal{N}\left( \lambda + \frac{\sigma_2^2}{\sigma_2^2 + \sigma_1^2}(v - \lambda), \frac{\sigma_2^2 \sigma_1^2}{\sigma_2^2 + \sigma_1^2} \right) \;.
    \label{eq:conditional_dis}
\end{align}

In the reverse process, the goal of the diffusion model is to find the estimator \( \hat{u} = f(v) \) that minimizes the MSE
\begin{align}
    \text{MSE} = \mathbb{E}\left[ \|u - \hat{u}\|_2^2 ~|~ v \right] = \int \|u - f(v)\|_2^2 p(u|v) \, du.
\end{align}

Taking the derivative of this expression with respect to \( f(v) \) and setting it to zero
\begin{align}
    \frac{d}{df(v)} \int \|u - f(v)\|_2^2 p(u|v) \, du \\
    = -2 \int (u - f(v)) p(u|v) \, du = 0 \;.
\end{align}

This simplifies to
\begin{align}
    \int u p(u|v) \, du = f(v) \int p(u|v) \, du = f(v) \;.
\end{align}

Thus, the closed-form solution for the minimum mean squared error (MMSE) estimator is
\begin{align}
    f_{\text{MMSE}}(v) = \mathbb{E}[u | v] \;.
\end{align}

From Eq.~\ref{eq:conditional_dis}, the explicit form of the MMSE estimator is
\begin{align}
    f_{\text{MMSE}}(v) = \lambda + \frac{\sigma_2^2}{\sigma_2^2 + \sigma_1^2}(v - \lambda) \;.
\end{align}

The variance of the MMSE solution is
\begin{equation}
  \scalebox{0.95}{$
    \begin{aligned}
       \text{Var}\left( \lambda + \frac{\sigma_2^2}{\sigma_2^2 + \sigma_1^2}(V - \lambda) \right) &= \left( \frac{\sigma_2^2}{\sigma_2^2 + \sigma_1^2} \right)^2 \text{Var}(V) \\
    &= \left( \frac{\sigma_2^2}{\sigma_2^2 + \sigma_1^2} \right)^2 \left( \text{Var}(U) + \sigma_1^2 \right) \\
    &=  \frac{\sigma_2^4}{\sigma_2^2 + \sigma_1^2} \;.
    \end{aligned}
  $}
\label{eq:proof_conclusion}
\end{equation}

\end{proof}

As discussed previously, in our context, the target distribution \( U \) that we aim to recover in the reverse process is the low-light noise distribution. For simplicity, we approximate it as a Gaussian distribution without sacrificing generality. Since the primary components of low-light noise (such as shot noise, read noise, and banding patterns) are either inherently modeled as Poisson distributions or can be approximated by Gaussian distributions~\cite{wei2021physics}. Moreover, with a sufficient photon count, a Poisson distribution can be approximated by a Gaussian, as a consequence of the central limit theorem.

\liying{More generally, as mentioned in the main paper, the same theory still holds for the Gaussian Mixture Models (GMMs). Assume the target distribution follows a GMM: $p(U) \sim \sum_{k=1}^K w_k \mathcal{N}(U; \mu_k, {\tau_k}^2)$, where $w_k$, $\mu_k$, ${\tau_k}^2$ are the mixture weights, means, and variances of the $k$-th Gaussian component, we have $p(U | V = v) = \sum_{k=1}^K \pi_k(v) \mathcal{N}\left(U; \mu_k + \frac{\tau_k^2}{\tau_k^2 + \sigma_1^2} (v - \mu_k), \frac{{\tau_k}^2 \sigma_1^2}{{\tau_k}^2 + \sigma_1^2}\right)$, where $\pi_k(v)$ is the posterior weight. Thus the variance of the MMSE solution is
$\sum_{k=1}^K \pi_k(v) \left( \frac{\tau_k^2}{\tau_k^2 + \sigma_1^2} \right)^2 \left( \sigma_1^2 + \text{Var}(u) \right)$.}

\begin{figure}[t]
  \centering
  \includegraphics[width=0.95\columnwidth]{ 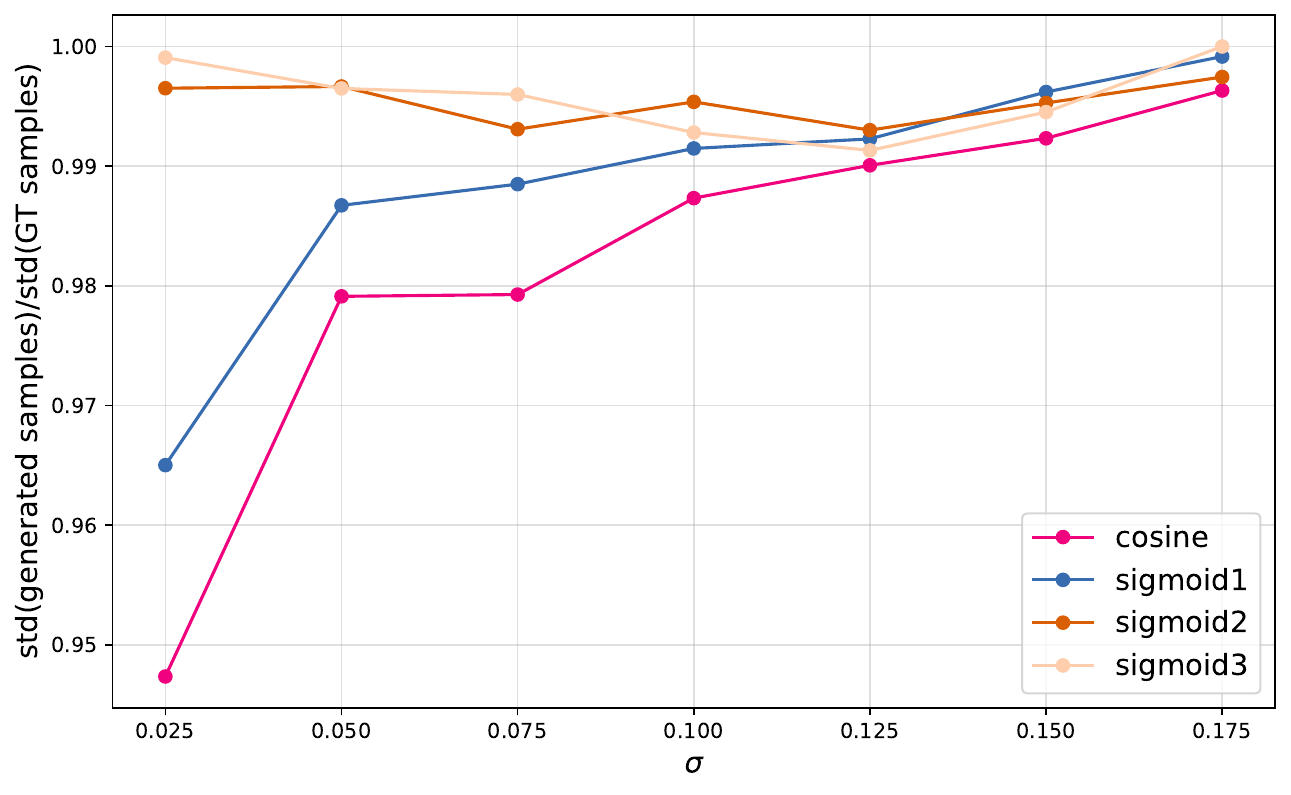}
  \caption{Experiments on seven TL distributions with different $\sigma$, where the x axis denotes the scale parameter $\sigma$ of these seven distributions, and the y axis shows the ratio of standard deviation between generated and GT samples. As can be observed, \textit{sigmoid3} (and \textit{sigmoid2}) produces standard deviations closer to the ground truth, while \textit{cosine} yields consistently lower stds}
  \label{fig:tukey_lambda}
\end{figure}

\liying{
\subsection{Experiment on 1D Tukey-Lambda Distributions}

We extend the experiment shown in Fig.~\ref{fig:effect_noise_scheduling} to a set of 1D Tukey-Lambda distributions, which have been shown in the ELD paper to closely approximate the distribution of read noise.
Specifically, we set the target distribution to Tukey-Lambda distributions with fixed shape ($\lambda=0.1$) and location ($\mu=0$) parameters while varying the scale parameter $\sigma$. We train a new 1D diffusion network to model seven different predefined distributions.

As shown in Fig.~\ref{fig:tukey_lambda}, for different Tukey-Lambda distributions, \textit{sigmoid3} and \textit{sigmoid2} produce standard deviations closer to the ground truth, whereas \textit{cosine} consistently yields lower standard deviations. This further validates our design. 

Additionally, as the scale parameter $\sigma$ increases, the performance gap between different noise schedules narrows. This is because, as indicated in Eq.\ref{eq:proof_conclusion}, when the signal variance $\sigma_2^2$ increases, the influence of the added Gaussian noise variance $\sigma_1^2$ diminishes.

We further conducted a statistical analysis of the noise standard deviation (std) distribution of the SID Sony dataset. For the highest ISO setting (25600) in the dataset, $80\%$ of the noise standard deviations fall within the range of $0.087\text{--}0.107$, and $99\%$ fall within the range of $0.087\text{--}0.173$. Within this range, selecting an appropriate noise schedule is crucial, as demonstrated in our experiments. For lower ISO settings, where the noise variances are smaller, the choice of noise schedule becomes even more significant.
}

\section{Implementation Details}
\label{supp:implementation}

This section provides additional information about the detailed network architecture, the datasets used for training and evaluation, the training configurations for our diffusion model and denoising network, and the process for generating synthetic pairs.

\subsection{Network Architecture}
\label{supp:architecture}

In this section, we show the details of the positional encoding and camera setting encoding for integrating pixel coordinate and camera setting information into the diffusion model.

\noindent \textbf{Positional Encoding.}
To integrate pixel coordinate information into our diffusion network, we propose to use positional encoding.
We first use sinusoidal positional encoding~\cite{vaswani2017attention} to encode the pixel coordinates as
\begin{align}
    \Tilde{\mathbf{c}} &= \mathrm{Conv}_{1\times 1}(\mathbf{c}) \;, \\
    \mathbf{p} &= \mathrm{Conv}_{1\times 1}([\Tilde{\mathbf{c}}, sin(\Tilde{\mathbf{c}}), cos(\Tilde{\mathbf{c}})]) \;,
\end{align}
where $\mathbf{c}$ is the coordinate map of the image pixels, $\mathbf{p}$ is the resulting positional embedding, $\mathrm{Conv}_{1\times 1}()$ denotes a convolutional layer with kernel size $1\times 1$, and $[~]$ denotes the concatenation in the channel dimension.

Subsequently, the positional embedding is integrated into the image feature $\mathbf{f}$ as
\begin{align}
    \mathbf{s}, \mathbf{r} &= \mathrm{Chunk}(\mathrm{Conv}_{1\times 1}(\mathbf{p})) \;, \\
    \Tilde{\mathbf{f}} &= \mathbf{f} \odot (1+\mathbf{s}) + \mathbf{r} \;,
\end{align}
where $\odot$ is the element-wise multiplication, and $\mathrm{Chunk}()$ represents the operation that splits the feature map into two chunks in the channel dimension with the same channel number. The learned features $\mathbf{s}$ and $\mathbf{r}$ are used to scale and shift the image feature.

\noindent \textbf{Camera Setting Encoding.}
Each camera setting's corresponding embedding $\mathbf{z}$ is integrated with the image feature $\mathbf{f}$ using cross-attention as
\begin{align}
    \mathbf{Q} &= \mathbf{f}\mathbf{W}_Q,~~ \mathbf{K}=\mathbf{z}\mathbf{W}_{K},~~ \mathbf{V}=\mathbf{z}\mathbf{W}_{V}  \;, \\
    \Tilde{\mathbf{f}} &= \mathbf{f} + Softmax(\frac{\mathbf{Q}\mathbf{K}^{T}}{\sqrt{d}})\mathbf{V} \;,
\end{align}
where $\mathbf{W}_Q$, $\mathbf{W}_K$ and $\mathbf{W}_V$ are convolution layers with $1\times 1$ kernels, and $d$ is the channel dimension of the $\mathbf{Q}$ and $\mathbf{K}$.

\subsection{Datasets}
\label{supp:datasets}

We use the SID Sony dataset to train and evaluate our network.
It comprises indoor and outdoor scenes captured using the Sony $\alpha$7S2 camera. 
The dataset comprises 1,865 images for training, 234 validation images, and 598 testing images, all with a resolution of $4240\times2832$ pixels. 
The noisy images are taken with short exposure times, whereas the corresponding clean images are captured with longer exposure times and are thus assumed to be noiseless.  
The exposure-time ratios between the long and short-exposure images are set at 100, 250, and 300. 

Additionally, we test our denoising network on the ELD Sony dataset~\cite{wei2021physics}; it consists of 60 low-light noisy raw images also captured using the same Sony $\alpha$7S2 camera model as the SID dataset.

\subsection{Training Configurations}
\label{supp:training}

Here we give more training details of our diffusion model and denoising network.
Note that our method operates in the RAW image domain, therefore, the Bayer arrays are packed into four channels~\cite{chen2018learning} (the spatial resolution is reduced by a factor of two in each dimension correspondingly) before being fed into diffusion or denosing networks.
This step is not shown in Fig.~2 of the main paper for clarity.

\noindent \textbf{Diffusion Model Training.}
Before training the diffusion model, we perform data resampling: for camera settings with fewer than 100 training pairs, we augment the data by repeating samples until the count reaches 100. 
We train our diffusion model using the Adam optimizer for 1,000 epochs. 
The learning rate is initially set to $1e-4$ and decays to $1e-5$ by the end of the training, using a cosine annealing schedule. 
The batch size is set to 12. 
We randomly crop $512\times 512$ patches from the training samples.
All Sony training pairs from the SID dataset are used for training, with their ISO values and exposure time ratios encoded.

After the training is complete, we use clean images in the SID training set to generate synthetic noisy images as described above.

\noindent \textbf{Denoising Network Training.}
The denoising networks are trained using the Adam optimizer for 500 epochs. During training, the batch size is 4, and we randomly crop $256 \times 256$ patches, applying random horizontal flips for data augmentation.

\subsection{Clean-Noisy Pair Generation}
\label{supp:pair_gen}

During diffusion model inference, we use clean images from the SID training set to generate synthetic data. 
As our diffusion network operates at an image size of $512 \times 512$, each clean image is first divided into overlapping patches of $512 \times 512$ pixels, with a 1/4 patch size overlap.
A pixel coordinate map is also computed for each patch.
We condition our network on these cropped patches, the corresponding pixel coordinate maps, and the camera settings to generate noise patches. 
Each noise patch is then added to its conditioned clean patch to create a noisy patch. 
These clean and noisy patches are paired together to form clean-noisy pairs subsequently used for training the denoising network.

\section{Evaluation Protocol: Noise Mean and Variance}
\label{sec:supp_evaluation_protocol}

In this section, we provide the detailed procedure for computing the noise mean and variance used for evaluating noise quality, 
as mentioned in Sec.~4.1 of the main paper.

\begin{figure}[t]
  \centering
  \includegraphics[width=0.7\columnwidth]{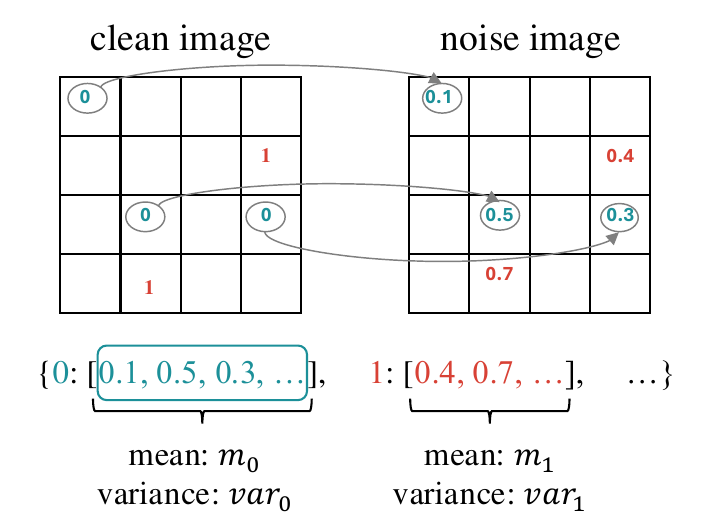}
  \caption{Noise mean and variance calculation. For a specific clean value $i$, we gather corresponding noise values from the noise image and compute their mean and variance. Here, we illustrate the clean values 0 and 1, for example. We perform the same procedure for all the clean values (e.g., for an 8-bit image, the clean values range from 0-255).}
  \label{fig:noise_mean_variance}
\end{figure}

The specific analysis procedure is indicated in Fig.~\ref{fig:noise_mean_variance}. 
Assuming the value range of the clean images falls within the interval $\left [ 0,M  \right ]$ (e.g., for an 8-bit image, $M=255$), we iterate over each clean pixel value. 
Given a clean pixel value $i$, all pixels in the clean images with this value are identified and their positions recorded. 
Subsequently, the pixels in the corresponding noise images at the same positions are retrieved. 
This process yields multiple noise values for clean pixel value $i$, from which the noise mean value $m_i$ and variance $var_i$ are computed. 
Some examples are illustrated in Fig.~5 of the main paper, 
where we compare these two statistics between real noise images and synthetic noise images. 
Correctly approaching these values is crucial for precisely training our denoising network.

\begin{figure*}[t]
  \centering
  \includegraphics[width=\linewidth]{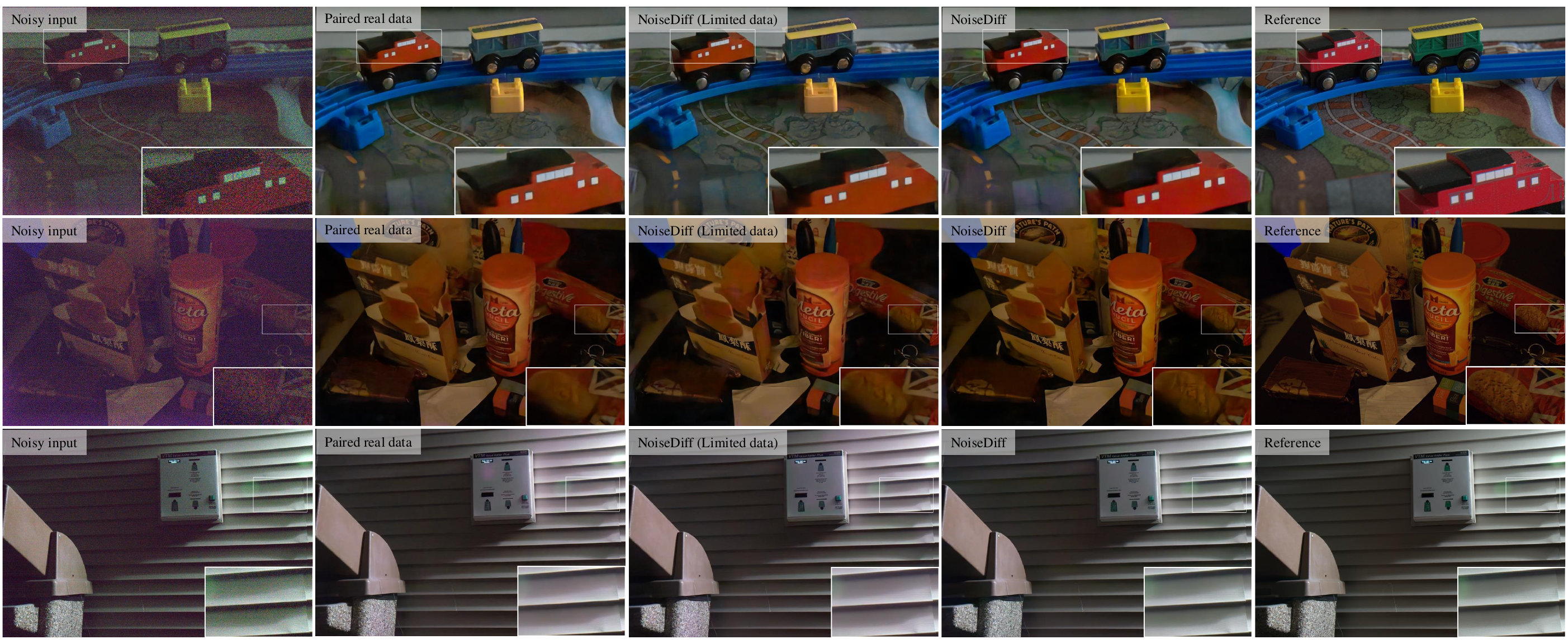}
  \caption{Ablation study on influence of training-data diversity for denoising. 
  \textbf{Paired real data:} The denoising network is trained using real data from the SID Sony training set~\cite{chen2018learning}, where some camera settings have very few unique scenes (e.g., the camera setting with ISO 6400 and ratio 300 has only one unique scenes).
  \textbf{NoiseDiff:} The denoising network is trained with our full set of synthetic data, where camera settings with limited unique scenes are augmented by scenes from other settings. Synthetic data is generated using the augmented scenes.
  \textbf{NoiseDiff (Limited data):} The denoising network is trained with limited synthetic data, where the number of scenes for each camera setting matches the original SID training set.
  The above three test samples from the SID testing set: (1) ISO 6400, ratio 300 (only one unique clean image in the original SID training set); (2) ISO 12800, ratio 100 (three unique clean images originally); and (3) ISO 800, ratio 100 (two unique clean images originally).
  It can be observed that the denoising performance improves with more synthetic data.}
  \label{fig:extraclean}
\end{figure*}

\begin{figure*}[t]
  \centering
  \includegraphics[width=1.0\linewidth]{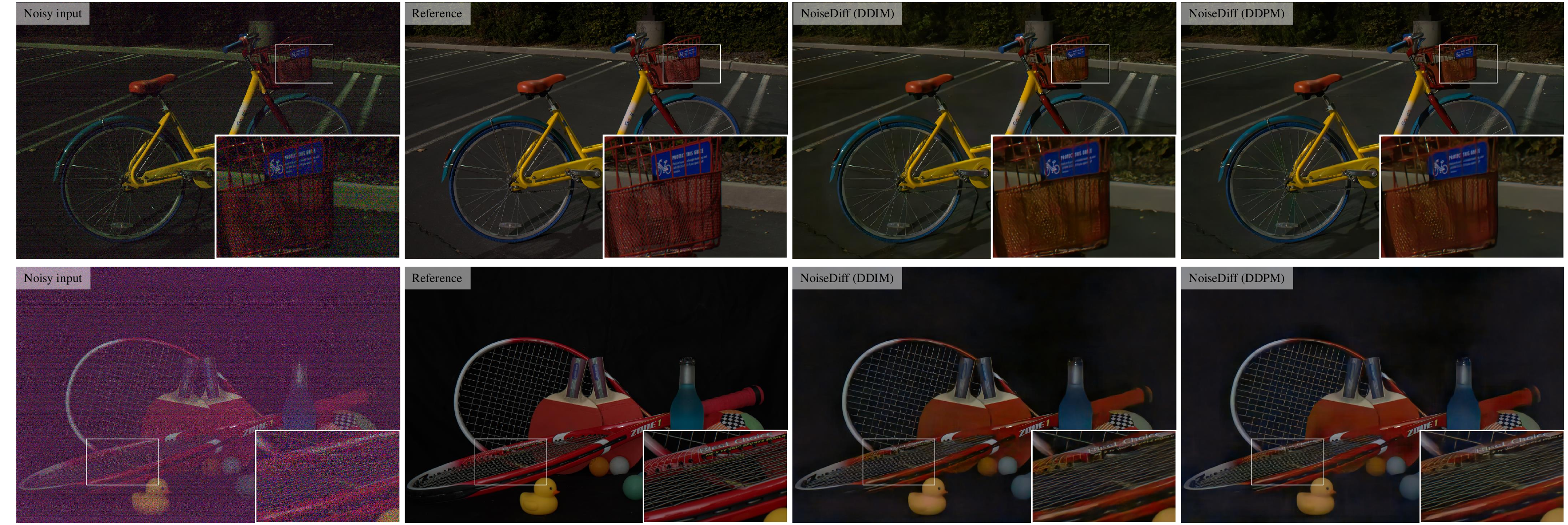}
  \caption{Comparison of denoising results from networks trained by synthetic data generated by DDIM~\cite{song2020denoising} and DDPM~\cite{ho2020denoising} sampling methods. We use 100 steps for DDIM sampling, and 1000 steps for DDPM sampling. It shows that DDIM can also generate visually satisfactory results, and can serve as an efficient alternative to our method.}
  \label{fig:ddpm_ddim_vis}
\end{figure*}

\section{More Discussions}
\label{supp:more_dis}

\subsection{Influence of Training-Data Diversity}
\label{supp:data_diversity}

In the main paper, we presented an ablation study investigating the influence of training data diversity. We trained a separate denoising network using less synthetic data than in our standard configuration. Here, we provide details about both our standard and limited-data configurations.

\noindent \textbf{Standard Configuration.}
The original SID training set contains varying numbers of scenes across different camera settings, with some settings having very few scenes. For instance, the camera setting with ISO 6400 and ratio 300 has only one unique scenes. With our data generation model, however, we can leverage all the clean images in the SID training set to generate numerous training pairs for each camera setting. Specifically, if there are $K$ clean images in total, we can generate $K$ noisy images for each camera setting by conditioning our diffusion model on the camera setting embedding. Furthermore, since DDPM sampling is non-deterministic, different random seeds allow us to generate infinite noisy images from a single clean image.
For efficiency, though, we generate a certain number of noisy images for each camera setting. For any setting with fewer than 6 unique clean images ($K_i < 6$), \textit{we randomly sample $30 - K_i$ additional clean images from other settings to augment the clean image number}, then generate noisy images from both the original and augmented clean images. For settings with more than 6 clean images, we use only the original clean images for data generation.

\noindent \textbf{Limited-Data Configuration.}
Here, we generate data without clean image augmentation. Specifically, for each camera setting with $K_i$ clean images in the SID training set, we generate noisy images using only these $K_i$ clean images. 
Therefore, the denoising network under this limited-data configuration uses the same number of scenes per camera setting as in the original SID training set.

As a result, these two settings consist of different number of synthetic pairs (only synthetic pairs, no real pairs included) for training the denoising networks.

Fig.~\ref{fig:extraclean} presents a visual comparison using three test samples from the SID testing set: (1) ISO 6400, ratio 300 (only one unique clean image in the original SID training set); (2) ISO 12800, ratio 100 (three unique clean images originally); and (3) ISO 800, ratio 100 (two unique clean images originally). As shown, \textit{NoiseDiff} produces better results than \textit{NoiseDiff (Limited data)}, benefiting from the additional synthetic training data. This further demonstrates that denoising performance improves with more synthetic data.

\begin{sidewaysfigure*}[htbp]
    \centering
    \setlength{\tabcolsep}{1pt}
    \includegraphics[width=1.0\linewidth]{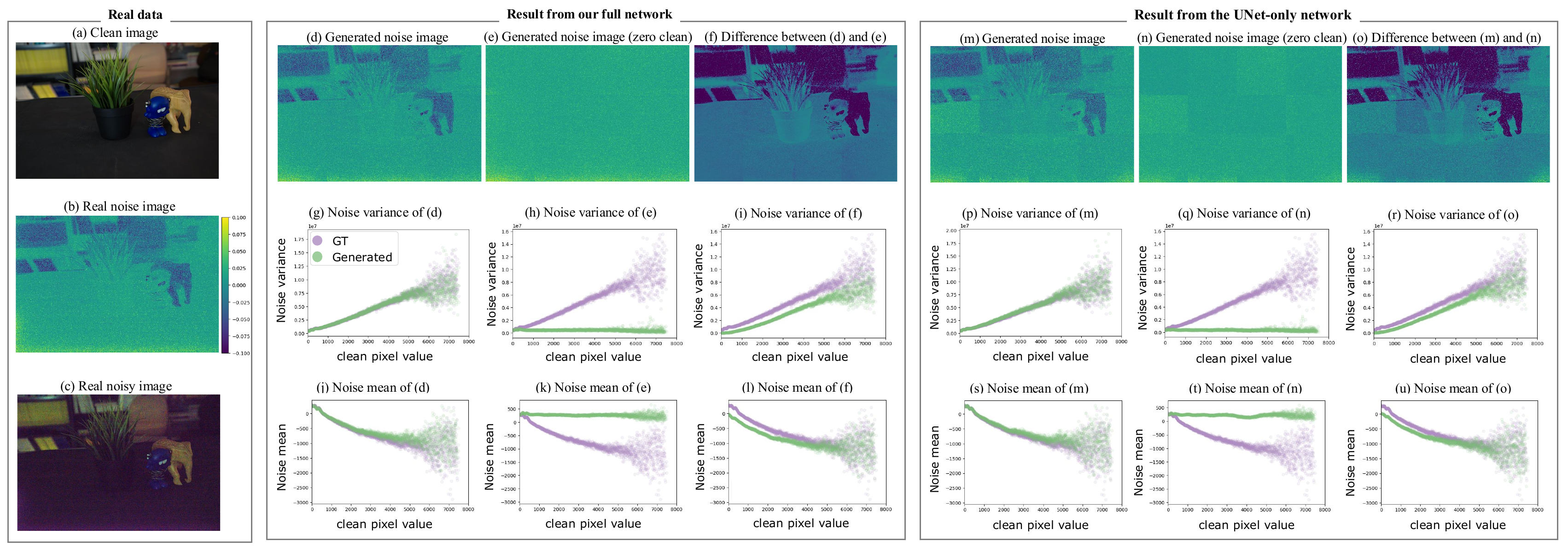}
    \caption{Noise decomposition of the generated noise. \textbf{(a)} A clean image from the SID Sony test set with ISO 4000 and ratio 300. \textbf{(b), (c)} The corresponding real noise image and noisy image. \textbf{(d)} The noise image generated by our full model. \textbf{(e)} When the clean image input, $\textbf{x}$, is set to a full zero map, the noise generated is the signal-independent noise component. \textbf{(f)} The signal-dependent noise component, obtained by subtracting (e) from (d). \textbf{(g), (j)} The noise variance and mean of (d). \textbf{(h), (k)} The noise variance and mean of (e). \textbf{(i), (l)} The noise variance and mean of (f). 
    \textbf{(m)-(u)} The results from the UNet-only network.
    For the visualized noise images, different noise values are represented by varying colors, with the color bar displayed on the right of (b).
    The statistics of the generated noise are shown in \textcolor[HTML]{7FBF7C}{green}, and the statistics of the real noise are shown in \textcolor[HTML]{B08DC3}{purple}. 
    The results indicate that our full model generates noise distributions closely resembling real noise.
    The mean and variance of the isolated signal-independent noise are independent of the clean pixel values. Conversely, the mean and variance of the signal-dependent noise are both correlated with the clean pixel values.
    While for the UNet-only network, it can be observed that the signal-independent and signal-dependent components are not fully disentangled, as shown in (m)-(o).
    }
    \label{fig:noise_decomposition}
\end{sidewaysfigure*}

\subsection{Sampling Method}
\label{supp:sampling}

\begin{figure*}[t]
  \centering
  \includegraphics[width=1.0\linewidth]{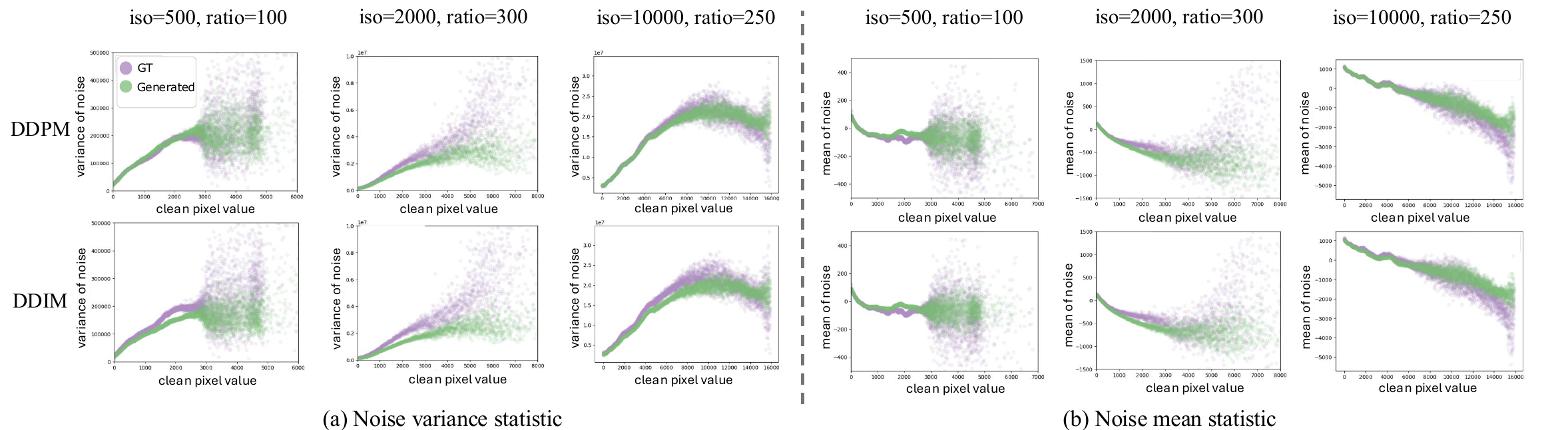}
  \caption{Comparison of noise statistics between DDIM, DDPM, and real noises across three different camera settings. The primary performance difference between the two sampling methods lies in the variance of the \textcolor[HTML]{7FBF7C}{generated noises}, with DDIM sampling exhibiting a larger variance gap from the \textcolor[HTML]{B08DC3}{real noise} compared to DDPM sampling.}
  \label{fig:ddpm_ddim}
\end{figure*}

As mentioned in the main paper, we use DDIM sampling as an efficient alternative for generating data.
Table~3 of the main paper shows that DDIM achieves satisfactory performance with a minor drop compared to DDPM-generated data. 
Qualitative results in Fig.~\ref{fig:ddpm_ddim_vis} further demonstrate that DDIM provides nearly comparable visual quality to DDPM, offering a practical alternative for faster generation.
To better understand the performance difference, we analyze noise statistics by computing the mean and variance. In Fig.~\ref{fig:ddpm_ddim}, we provide comparisons of the noise statistics between DDPM and DDIM. As can be observed, DDIM sampling tends to yield low-light noise with lower variance.

\subsection{Noise Decomposition}
\label{supp:noise_decomp}


\noindent \textbf{Implicit Noise Decomposition.}
While noise decomposition on real noise images is not possible due to the absence of ground-truth for the signal-dependent and signal-independent components, the conditional generation capability of our diffusion model allows us to perform implicit noise decomposition on the generated noise.

In Fig.~\ref{fig:noise_decomposition} (a), (b), and (d)-(f), we visualize the clean image, the ground-truth noise, the noise generated by our full model, and the decomposed components. 
To achieve this decomposition, we condition the diffusion model on a fully black clean image (a zero map), changing the generation process from $\Tilde{\mathbf{n}}_{t} = f_{\theta} (\Tilde{\mathbf{n}}_{t-1}, t, \mathbf{x}, \mathbf{c}, \mathbf{z})$ (as described in Sec.~3.1 of the main paper) to $\Tilde{\mathbf{n}}_{t} = f_{\theta} (\Tilde{\mathbf{n}}_{t-1}, t, \mathbf{0}, \mathbf{c}, \mathbf{z})$. This modification removes the signal information, resulting in the generation of only the signal-independent noise component, shown in Fig.~\ref{fig:noise_decomposition} (e). Subtracting the signal-independent noise from the full noise isolates the signal-dependent noise component, as shown in Fig.~\ref{fig:noise_decomposition} (f). Notably, the row noise and low-frequency noise are captured in Fig.~\ref{fig:noise_decomposition} (e), whereas Fig.~\ref{fig:noise_decomposition} (f) contains primarily a shot-noise-like pattern.

We further analyze the statistics of different noise components. In Fig.~\ref{fig:noise_decomposition} (g) and (j), we compare the variance and mean values of our generated noise with those of the real noise; they are calculated using the method described in Sec.~\ref{sec:supp_evaluation_protocol}. The results indicate that our method produces a noise distribution similar to the real noise. In Fig.~\ref{fig:noise_decomposition} (h) and (k), we present the statistics of the signal-independent noise; they clearly show that the noise mean and variance are independent of the clean pixel values. Conversely, Fig.~\ref{fig:noise_decomposition} (i) and (l) depict the statistics of the signal-dependent noise, demonstrating that both the noise variance and mean are correlated with the clean pixel values.

This noise decomposition capability allows our diffusion model to generate either full noise maps or only signal-independent noise maps (dark frames). Dark frames are valuable for computing dark shadings, as discussed in Sec.~2 of the main paper, which can further aid in the denoising process. 

Drawing inspiration from \cite{feng2022learnability}, this noise decomposition could be further used for faster full noise generation by using our diffusion model to generate dark frames and a fitted Poisson distribution for signal-dependent noise. Generating only a few dark frames with the diffusion model would allow for on-the-fly noise generation, which is impossible due to the lengthy diffusion process.

\noindent \textbf{Additional Evidence Supporting the Effectiveness of the Two-Branch Network.}
Meanwhile, we also apply the same decomposition process to the noise generated by the UNet-only architecture, with the results shown in Fig.\ref{fig:noise_decomposition} (m)-(u). 
It is evident that the signal-independent and signal-dependent components are not fully disentangled. 
For instance, in Fig.\ref{fig:noise_decomposition} (o), low-frequency noise is still present at the bottom of the signal-dependent component.
This performance difference between the two architectures further highlights the effectiveness of the two-branch architecture in modeling signal-independent and signal-dependent noise, simplifying the network’s learning process.

\begin{table}[t]
    \caption{Ablation study on the dark shading correction (DSC) and shot noise augmentation (SNA) methods. We show the PSNR/SSIM results on the SID and ELD Sony test sets.}
    \centering
    \begin{tabular}{llccc}
        \toprule
        \multicolumn{2}{l}{} & Model 1 & Model 2 & Model 3 \\
        \midrule
        \multirow{2}{*}{Methods} 
        & DSC & \xmark & \cmark & \cmark \\
        & SNA & \xmark & \xmark & \cmark \\
        \midrule
        \multirow{3}{*}{SID} 
        & $\times 100$ & 43.30 / 0.958 & 43.76 / \best{0.961} & \best{43.92} / \best{0.961} \\
        & $\times 250$ & 40.53 / 0.944 & 41.26 / 0.945 & \best{41.28} / \best{0.946} \\
        & $\times 300$ & 37.68 / 0.928 & 37.89 / \best{0.930} & \best{37.90} / 0.929 \\
        \midrule
        \multirow{2}{*}{ELD} 
        & $\times 100$ & 45.79 / 0.972 & 46.51 / \best{0.979} & \best{46.95} / 0.978 \\
        & $\times 200$ & 42.25 / 0.924 & 44.90 / \best{0.972} & \best{45.11} / 0.971 \\
        \bottomrule
    \end{tabular}
    
    \label{tab:dsc_sna}
\end{table}

\begin{figure*}[htp]
  \centering
  \includegraphics[width=0.85\linewidth]{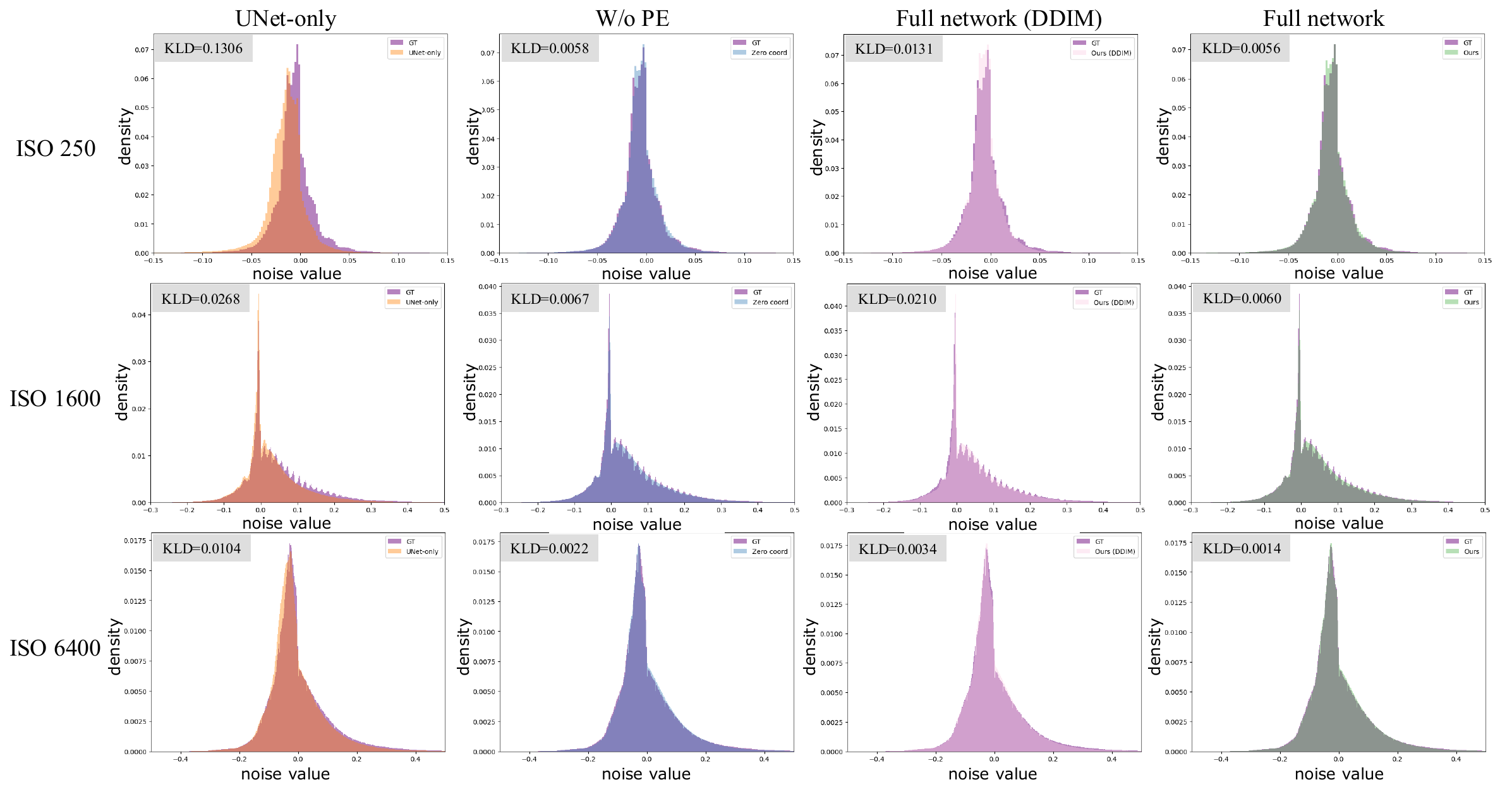}
  \caption{Histogram of noise values and Kullback-Leibler (KL) divergence between generated noise and \textcolor[HTML]{984ea3}{real noise} for four ablation variants: the \textcolor[HTML]{ff7f00}{UNet-only} architecture, model without positional encoding (\textcolor[HTML]{377eb8}{W/o PE}), our full model with DDIM sampling (\textcolor[HTML]{FDEBF4}{Full network (DDIM)}) and our full model with DDPM sampling (\textcolor[HTML]{4daf4a}{Full network}). Evaluations are conducted across three different ISO settings. The results produce by our full model are closest to the real noise distribution.}
  \label{fig:kld}
\end{figure*}

\subsection{Noise Histogram}
\label{supp:noise_hist}

Here, we present the histogram of generated noise from different ablation variants. As shown in Fig.~\ref{fig:kld}, our full model with DDPM sampling generates noise that aligns more closely with the distribution of real noise.

\liying{\subsection{Influenc of Dark Shading Correction and Shot Noise Augmentation}
\label{supp:dsc_sna}

We employ dark shading correction (DSC) and shot noise augmentation (SNA) when training the denoising network of \textit{NoiseDiff*} to ensure a fair comparison with LRD and PMN. 
As shown in Table~\ref{tab:dsc_sna}, with the additional dark shading data, DSC significantly enhances denoising performance. SNA yields a smaller improvement. The combination of DSC and SNA achieves the best overall performance.
}

\section{More Visual Comparisons}
\label{supp:more_visual}

\begin{figure*}[t]
  \centering
  \includegraphics[width=\linewidth]{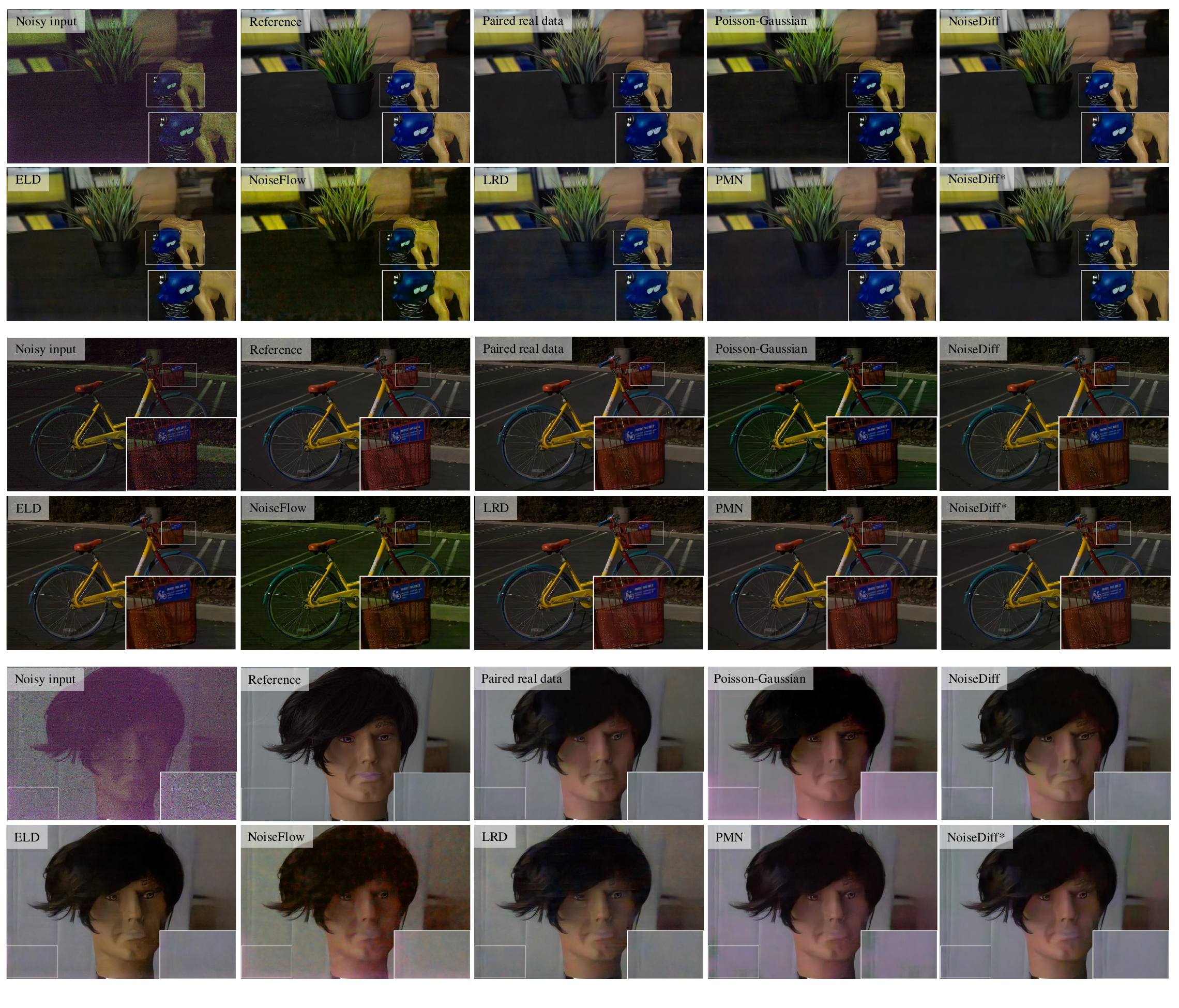}
  \caption{Visual comparisons of denoising results from networks trained on data synthesized using different methods, alongside comparisons with a model trained on real clean-noisy pairs from the SID training set. All test samples are from the SID Sony test set. Zoom in for a clearer view.}
  \label{fig:comparison_sid_supp1}
\end{figure*}

\begin{figure*}[t]
  \centering
  \includegraphics[width=\linewidth]{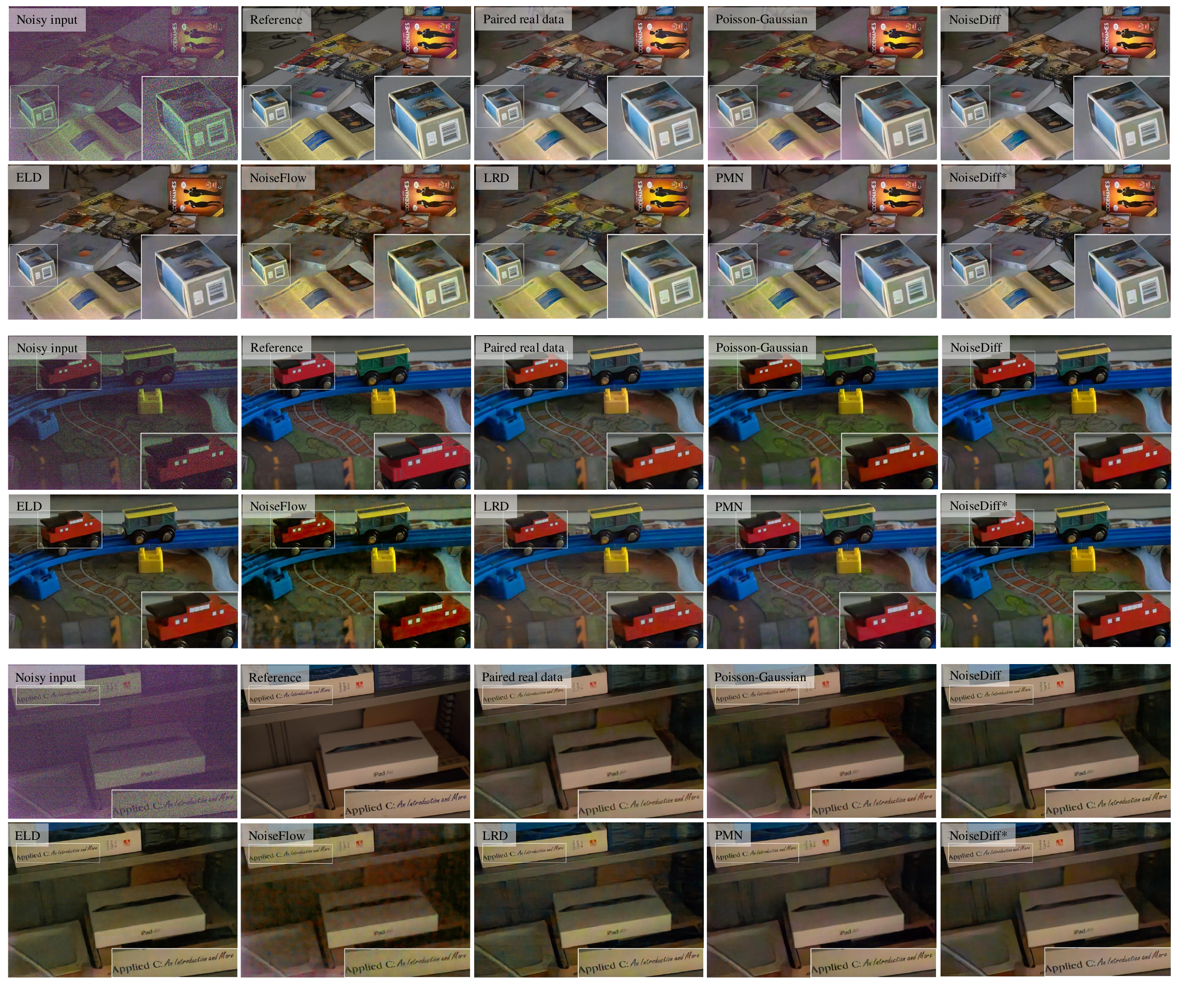}
  \caption{Visual comparisons of denoising results from networks trained on data synthesized using different methods, alongside comparisons with a model trained on real clean-noisy pairs from the SID training set. All test samples are from the SID Sony test set. Zoom in for a clearer view.}
  \label{fig:comparison_sid_supp2}
\end{figure*}

\begin{figure*}[t]
  \centering
  \includegraphics[width=\linewidth]{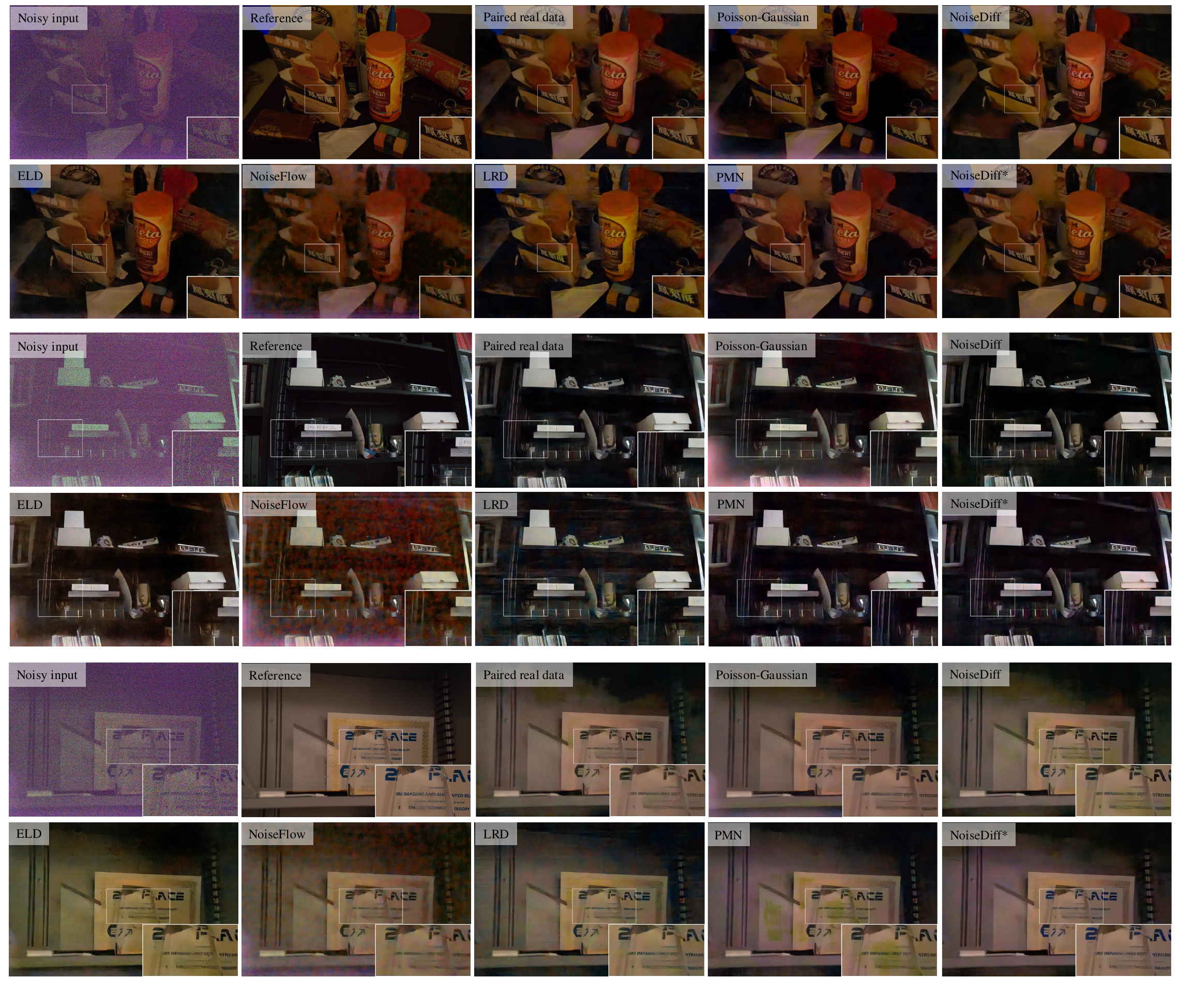}
  \caption{Visual comparisons of denoising results from networks trained on data synthesized using different methods, alongside comparisons with a model trained on real clean-noisy pairs from the SID training set. All test samples are from the SID Sony test set. Zoom in for a clearer view.}
  \label{fig:comparison_sid_supp3}
\end{figure*}

\begin{figure*}[t]
  \centering
  \includegraphics[width=\linewidth]{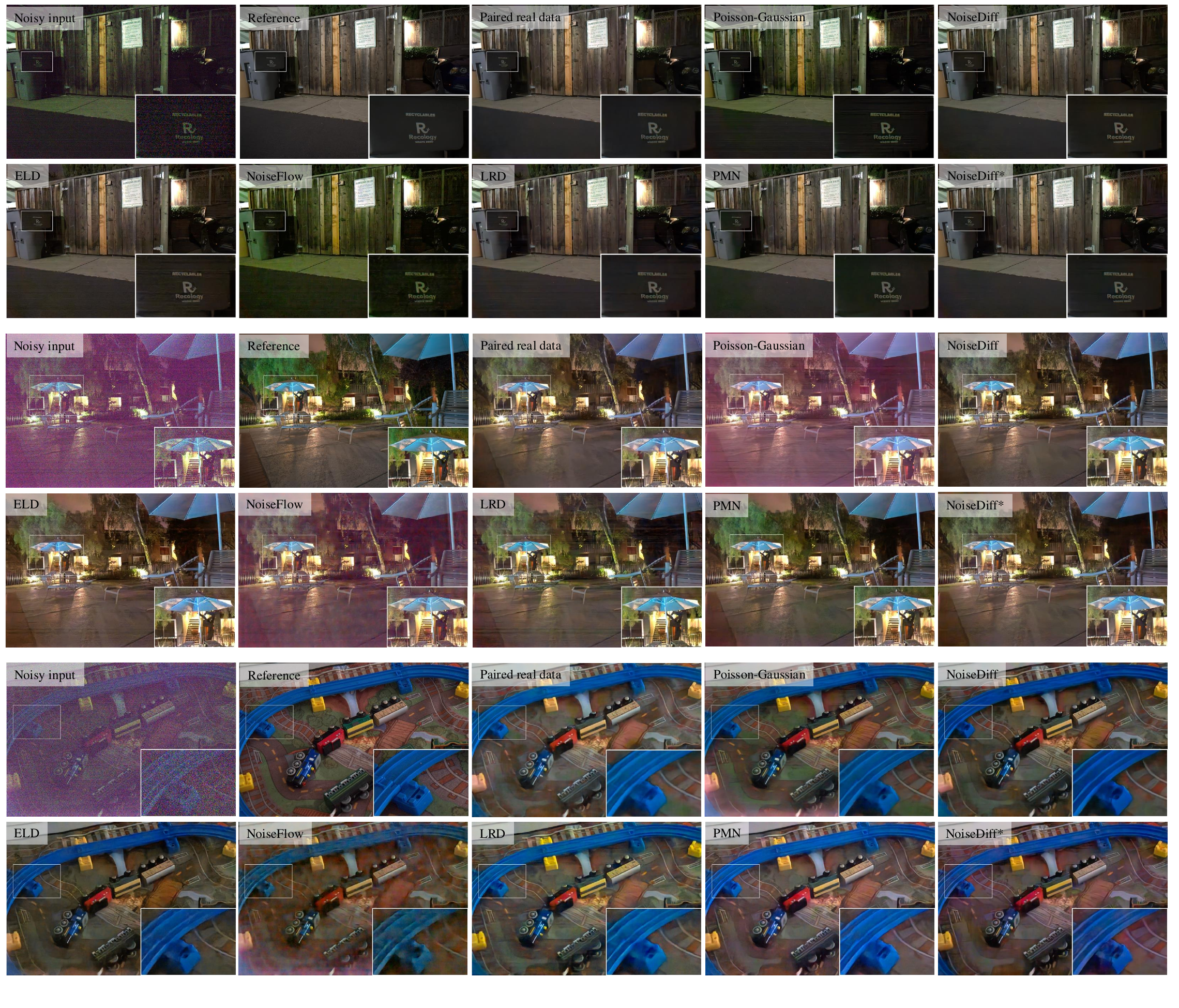}
  \caption{Visual comparisons of denoising results from networks trained on data synthesized using different methods, alongside comparisons with a model trained on real clean-noisy pairs from the SID training set. All test samples are from the SID Sony test set. Zoom in for a clearer view.}
  \label{fig:comparison_sid_supp4}
\end{figure*}

\begin{figure*}[t]
  \centering
  \includegraphics[width=\linewidth]{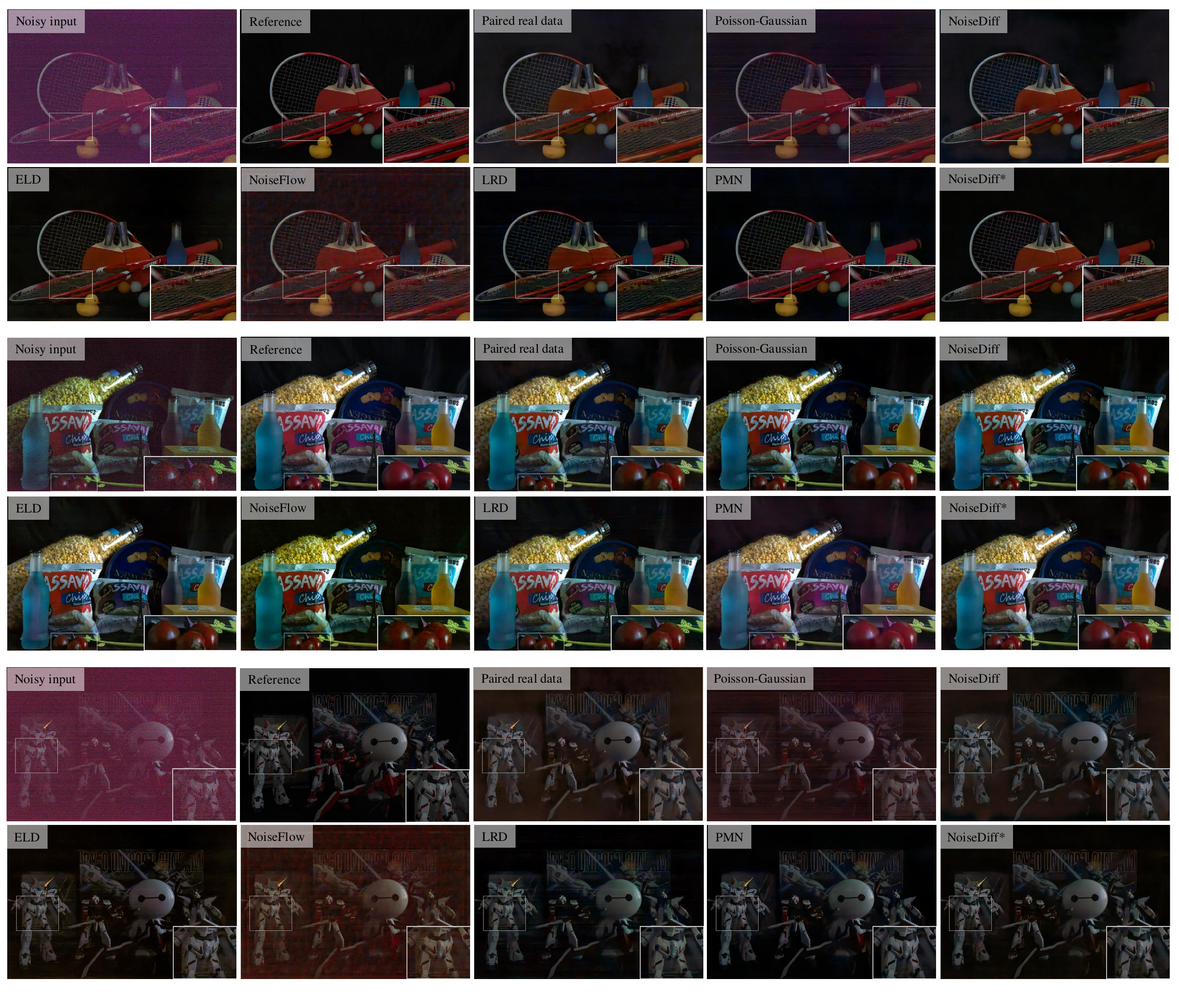}
  \caption{Visual comparisons of denoising results from networks trained on data synthesized using different methods, alongside comparisons with a model trained on real clean-noisy pairs from the SID training set. All test samples are from the ELD test set. Zoom in for a clearer view.}
  \label{fig:comparison_eld_supp1}
\end{figure*}

\begin{figure*}[t]
  \centering
  \includegraphics[width=\linewidth]{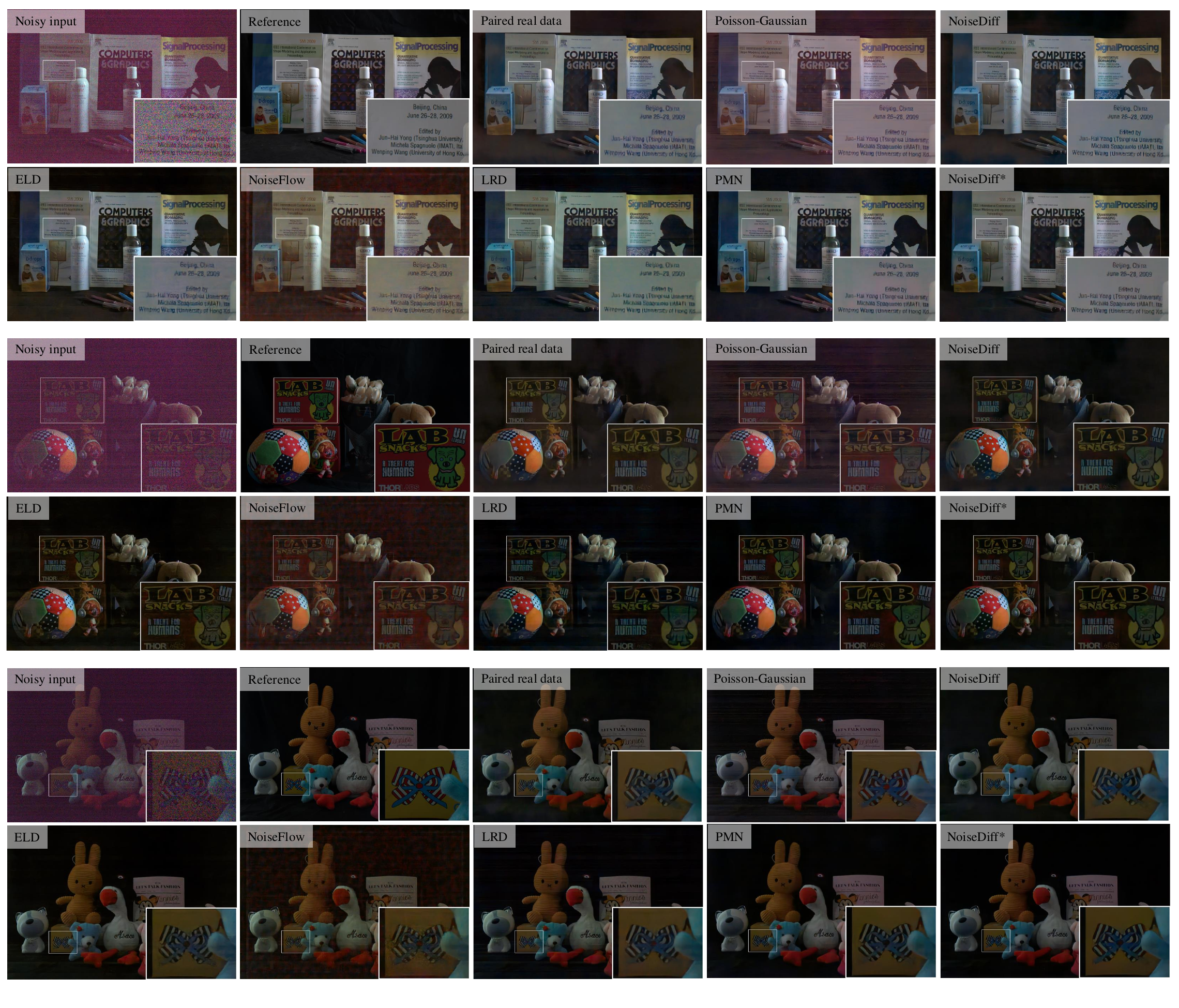}
  \caption{Visual comparisons of denoising results from networks trained on data synthesized using different methods, alongside comparisons with a model trained on real clean-noisy pairs from the SID training set. All test samples are from the ELD test set. Zoom in for a clearer view.}
  \label{fig:comparison_eld_supp2}
\end{figure*}

In this section, we provide more visual comparisons of denoising results from different networks trained on data synthesized using different methods. Fig.~\ref{fig:comparison_sid_supp1} - Fig.~\ref{fig:comparison_sid_supp4} are examples from the SID test set and Fig.~\ref{fig:comparison_eld_supp1} - Fig.~\ref{fig:comparison_eld_supp2} are examples from the ELD test set.

Our method excels at mitigating low-frequency noise, particularly visible in the third example of Fig.~\ref{fig:comparison_sid_supp1}, the first examples of Fig.~\ref{fig:comparison_sid_supp2} and the second example of Fig.~\ref{fig:comparison_sid_supp3}. Moreover, it enhances detail retention in denoising results, for example, producing clearer text compared to other methods, as shown in the third examples of Fig.~\ref{fig:comparison_sid_supp2}, the first example of Fig.~\ref{fig:comparison_sid_supp3} and the first example of Fig.~\ref{fig:comparison_eld_supp2}.
Additionally, our method demonstrates effectiveness in handling row noise, as evidenced in the third examples of Fig.~\ref{fig:comparison_eld_supp1}.

\end{document}